\documentclass{article}
\usepackage{microtype}
\usepackage{graphicx}
\usepackage{natbib}
\usepackage[accepted]{icml2022}

\usepackage{mathtools} %
\usepackage{amsthm}
\usepackage{centernot}
\usepackage{thmtools}
\usepackage{thm-restate}
\usepackage{nicefrac}       
\usepackage{wrapfig}
\usepackage{subcaption}
\usepackage{booktabs} %
\usepackage{adjustbox}
\usepackage{caption}
\usepackage{tikz}
\usepackage{comment}
\usepackage[toc,page]{appendix}
\usetikzlibrary{shapes.geometric,positioning,bayesnet,calc,intersections,through,backgrounds}
\usetikzlibrary{arrows,shapes,plotmarks,positioning}
\usetikzlibrary{decorations.markings}
\usepackage{tkz-euclide}
\usepackage{pgfplots}
\tikzset{>=stealth'} 
\tikzstyle{graphnode} = 
   [circle,draw=black,minimum size=22pt,text centered,text
     width=22pt,inner sep=0pt] 
\tikzstyle{var}   =[graphnode,fill=white]
\tikzstyle{vardashed}   =[graphnode,draw=gray,fill=white]
\tikzstyle{obs}   =[graphnode,fill=black,text=white]
\tikzstyle{obsgrey}   =[graphnode,draw=white,fill=lightgray,text=black]
\tikzstyle{par}    =[graphnode,draw=white,fill=red,text=black] 
 \tikzstyle{crucial} =[graphnode,draw=white,fill=yellow,text=black] 
\tikzstyle{fac}   =[rectangle,draw=black,fill=black!25,minimum size=5pt]
\tikzstyle{facprior} =[rectangle,draw=black,fill=black,text=white,minimum size=5pt]
\tikzstyle{edge}  =[draw=white,double=black,very thick,-]
\tikzstyle{blueedge}  =[draw=white,double=blue,very thick,-]
\tikzstyle{rededge}  =[draw=white,double=red,very thick,-]
\tikzstyle{prior} =[rectangle, draw=black, fill=black, minimum size=
5pt, inner sep=0pt]
\tikzstyle{dirprior} = [circle, draw=black, fill=black, minimum
size=5pt, inner sep=0pt]
\tikzstyle{dot_node}=[draw=black,fill=black,shape=circle]
\usepackage{colour-blind}
\definecolor{blue_cblind}{HTML}{1A85FF}
\definecolor{red_cblind}{HTML}{D41159}

\usepackage[colorlinks=true, linkcolor=red, urlcolor=blue, citecolor=blue, anchorcolor=blue]{hyperref}
\usepackage{url}
\usepackage{cleveref}
\crefname{figure}{Fig.}{Figs.}
\crefname{definition}{Defn.}{Defns.}
\crefname{corollary}{Corollary}{Corollaries}
\crefname{lemma}{Lemma}{Lemmas}
\crefname{proposition}{Prop.}{Props.}
\crefname{theorem}{Thm.}{Thms.}
\crefname{assumption}{Asm.}{Asms.}
\crefname{remark}{Remark}{Remarks}
\crefname{principle}{Principle}{Principles}
\crefname{lemma}{Lemma}{Lemmas}
\crefname{table}{Tab.}{Tabs.}
\crefname{section}{\S}{\S\S}
\crefname{subsection}{\S}{\S\S}
\crefname{subsubsection}{\S}{\S\S}
\crefname{appendix}{Appx.}{Appxs.}
\usepackage{enumitem}
\setlist[itemize]{leftmargin=*,itemsep=0em,topsep=0em}
\setlist[enumerate]{leftmargin=*,itemsep=0em,topsep=0em}
\usepackage{csquotes}
\usepackage{soul}
\usepackage{tcolorbox}
\usepackage{notation}
\usepackage[normalem]{ulem}
\usepackage[super]{nth}
\usepackage{nccmath}
\usetikzlibrary{patterns}
\usepackage[abs]{overpic}

\usepackage{diagbox}

\usepackage{nicematrix}

\newcommand{\elkeki}[1]{\textcolor{blue}{[#1]}}
\newcommand{\luigi}[1]{\textcolor{red}{[\textbf{Luigi:} #1]}}
\newcommand{\dominik}[1]{\textcolor{orange}{[\textbf{Dominik:} #1]}}
\newcommand{\jonas}[1]{\textcolor{cb-green-sea}{[\textbf{Jonas:} #1]}}
\newcommand{\bernhard}[1]{\textcolor{blue}{{\bf B: #1}}}
\newcommand{\julius}[1]{\textcolor{purple}{{\textbf{Julius:} #1}}}
\renewcommand{\elkeki}[1]{}
\renewcommand{\luigi}[1]{}
\renewcommand{\dominik}[1]{}
\renewcommand{\julius}[1]{}
\renewcommand{\jonas}[1]{}
\renewcommand{\bernhard}[1]{} %

\newcommand{\negspace}{-0.em}
\newcommand{\negspaceeq}{-0.em}

\icmltitlerunning{Causal Inference through the Structural Causal Marginal Problem}

\begin{document}

\twocolumn[
\icmltitle{Causal Inference Through the Structural Causal Marginal Problem
}

\icmlsetsymbol{equal}{*}

\begin{icmlauthorlist}
\icmlauthor{Luigi Gresele}{equal,mpi}%
\icmlauthor{Julius von Kügelgen}{equal,mpi,cambridge}
\icmlauthor{Jonas M. Kübler}{equal,mpi}%
\icmlauthor{Elke Kirschbaum}{ama}
\icmlauthor{Bernhard Schölkopf}{mpi}
\icmlauthor{Dominik Janzing}{ama}
\end{icmlauthorlist}

\icmlaffiliation{mpi}{Max Planck Institute for Intelligent Systems, Tübingen, Germany}
\icmlaffiliation{cambridge}{University of Cambridge, Cambridge, United Kingdom}
\icmlaffiliation{ama}{Amazon Research, Tübingen, Germany}

\icmlcorrespondingauthor{Luigi Gresele}{luigi.gresele@tue.mpg.de}
\icmlkeywords{Machine Learning, ICML}

\vskip 0.3in
]

\printAffiliationsAndNotice{\icmlEqualContribution} %
\begin{abstract}
We introduce an approach to counterfactual inference based on merging information from multiple datasets.
We consider a causal reformulation of the statistical \textit{marginal problem}: given a collection of {\em marginal} structural causal models (SCMs) over distinct but overlapping sets of variables, determine the set of {\em joint} SCMs that are counterfactually consistent with the marginal ones.
We formalise this approach for categorical SCMs using the response function formulation and show that 
it reduces the space of allowed marginal and joint SCMs.
Our work thus highlights a new mode of falsifiability through 
additional \textit{variables},
in contrast to the statistical one via additional \textit{data}.

%

%

%

%
%
%
%
%
%
%
%
%
%
%
\end{abstract}

\section{Introduction}
\label{sec:intro}
Counterfactual statements are ubiquitous in human judgement and reasoning. 
Consider the following example. %
A patient, Alice, is recommended a treatment $X$ against her disease and agrees to take it. The effectiveness of the treatment has been rigorously established through a randomised control trial, %
which found a positive average causal effect (ACE). %
However, the ACE is an average of treatment efficacy over the whole population, including some individuals who respond better and others who respond worse. %
Alice might wonder %
what {\em her own} chances of recovery would have been, had she not taken $X$---%
a query called the effect of treatment on the treated (ETT)~\cite{heckman1992randomization,shpitser2009effects}.
This requires envisioning consequences of %
a hypothetical change (not taking the treatment), given that the opposite happened (in reality, she took it). 
In a proposed hierarchy of causal reasoning termed the \textit{ladder of causation}~\citep{pearl2018book, bareinboim2020pearl}, such counterfactual statements occupy the highest, third rung, whereas the second rung corresponds to interventions and experiments~(``doing'') and the lowest, first rung to passive observation~(``seeing'').
Counterfactual reasoning (e.g., answering personalised, individual-level questions such as Alice's) thus requires the 
most fine-grained causal modelling.\footnote{the example concerns an individual causal effect; population-level counterfactuals can also be considered~\citep[][\S~3.4]{pearl2009causal}.}
In the 
graphical 
approach to causal inference~\cite{pearl2009causality}, counterfactuals are expressed using structural causal models (SCMs).

\newcommand{\yshift}{-3em}
\newcommand{\xshift}{5em}
\newcommand{\xzcol}{cb-burgundy!60!white}
\newcommand{\yzcol}{cb-blue!60!white}
\begin{figure}
\vspace{\negspace}
\vspace{\negspace}
    \centering
    \begin{tikzpicture}
		\centering
		\node (X) [obs,thick] {$X$};
		\node (Y) [obs, xshift=2*\xshift] {$Y$};
		\node (Z) [obs, xshift=\xshift, yshift=\yshift] {$Z$};
		\plate[yshift=-0.em,
		inner sep=0.5em, 
		ultra thick,
		color=\yzcol] {M2}{(Y) (Z)}{\textcolor{\yzcol}{\textbf{Dataset 2}}}; 
		\tikzset{plate caption/.style={caption, node distance=0, inner sep=0pt, below left=5pt and 0pt of #1.south,text height=.2em,text depth=0.3em}}
		\plate[yshift=0.3em,
		inner sep=0.5em,
		ultra thick,
		color=\xzcol] {M1}{(X) (Z)}{\textcolor{\xzcol}{\textbf{Dataset 1}}};
		\edge [color=\xzcol, thick] {X}{Z};
		\edge [color=\yzcol, thick] {Y}{Z};
	\end{tikzpicture}
	\vspace{\negspace}
    \caption{%
    \textbf{Overview of the Causal Marginal Problem.}
    Given observations of subsets of variables in a causal graph, consistently merging the available \textit{causal} marginal information imposes non-trivial constraints on the set of admissible joint and marginal 
    causal models which can in turn be useful for counterfactual inference.
    }
    \vspace{\negspace}
	\vspace{\negspace}
    \label{fig:teaser}
\end{figure}
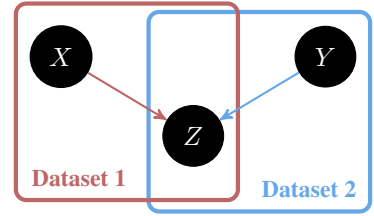
In practice, however,
we typically do not have access to an SCM 
but only to observational or experimental data (rungs one and two)
which may be insufficient to answer questions such as Alice's: we simply cannot perform an experiment where the same person is both given and not given a treatment,
an issue also referred to as the {\em fundamental problem of causal inference}~\cite{Imbens2015}. 
Counterfactual queries thus need to be evaluated based on a partial state of knowledge and
may be subject to an unresolvable degree of ambiguity, even in the absence of statistical uncertainty~\cite{dawid2000causal}.
\citet{pearl2011logic} therefore postulates restrictions on the types of 
inference we can make
given our data and modelling assumptions: %
counterfactual expressions should be evaluated subject to an {\em identifiability} requirement, specifying whether a given query can be estimated based on empirical observations, under conditions which can be phrased in the language of graphical models~\citep{shpitser2007counterfactuals, shpitser2008complete, Pearl_indirect, correa2021nested}. 

When full identification is not achievable, partial identification
sometimes still yields
informative bounds 
based on empirically observable quantities~\citep{robins1989analysis,
manski1990nonparametric,
balke1997bounds, tian2000probabilities}.
However, these methods typically rely on 
{\em  joint information} over all variables, based on observational or experimental studies, or combinations thereof~\citep{zhang2021partial}.

\textit{What if we instead have studies involving distinct, but overlapping subsets of variables? Can we combine them %
to answer counterfactual questions?}
In Alice's case, knowing the effect of treatment~$X$ alone may be insufficient. %
Suppose, however, that a separate study %
characterises the interventional effect of a rare condition $Y$ on her disease~(cf.~\cref{fig:teaser}). 
Since the condition is rare, and testing for it is costly, there are no studies characterising the joint effect of~$X$ and~$Y$ on recovery. %
\looseness-1 Could Alice nevertheless make use of the available information %
on the effect of $Y$ and combine it with information on $X$ to better answer her counterfactual question?
In order to answer these kinds of questions, in the present work %
we propose an approach to counterfactual causal inference %
which does not %
require joint observations of all variables: instead, our approach is
based on \emph{merging information from different datasets}, involving distinct but overlapping sets of variables.
This can be seen as the \textit{causal reformulation} of a classic problem in statistics called the \textit{marginal problem}~\citep{vorob1962consistent, Kellerer1964}.

\begin{tcolorbox}[boxsep=-0.25em]
\textbf{(Statistical) Marginal Problem:}
\textit{Given some distributions over non-identical but overlapping subsets of variables, determine existence and uniqueness of a consistent joint distribution over their union.}
\end{tcolorbox}
For example, consider random variables $X$, $Y$, $Z$, and suppose that we are given the {\em ``marginals''}  $\PP_{XZ}$ and $\PP_{YZ}$.
Is there a joint $\PP_{XYZ}$ that implies these marginals?\footnote{A trivial negative example is the case where $\PP_{XZ}$ and $\PP_{YZ}$ imply different $\PP_Z$; in general, $\PP_{XYZ}$ (if it exists) is not unique.}

In our proposed causal reformulation, we aim to merge  \textit{marginal causal models} such that they are consistent at various levels of the ladder of causation. 
In particular, we focus on the \emph{counterfactual marginal problem}, in which counterfactual consistency across marginal and joint SCMs is enforced. %
We formalise this in the context of categorical SCMs by exploiting their  \emph{response function formulation}~\citep{greenland1986identifiability, balke1994counterfactual}
 and show that
counterfactuals can %
acquire empirical content when considered in the broader context of a joint model, %
{\em even if only observations of the marginal models 
are available}.

\paragraph{Structure and contributions.} %
Following a review of relevant notions of causal modelling%
~(\cref{sec:background}),  we introduce the structural causal marginal problem~(\cref{sec:counterfactual_marginal_problem}), describe how to treat it~(\cref{subsec:solutions}) and illustrate its applications through examples~(\cref{sec:examples}), theory~(\cref{sec:theory}) and numerical simulations~(\cref{sec:experiments}).
Finally, we describe extensions of the basic setting~(\cref{sec:extensions}) and discuss our findings in the context of existing literature~(\cref{sec:discussion}).
%

While focusing mostly on simple examples,
the present work still makes a significant conceptual point: 
SCMs can sometimes be falsified as interventional models over additional
{\em variables} become available. This provides causal models with an additional mode of falsifiability compared to
statistical models, where the standard is to do this by means of additional {\em data}. %
The boundaries between the first two rungs of the ladder of causation %
and the third thus %
become more blurry as additional variables are observed.

\section{Categorical Structural Causal Models}
\label{sec:background}
An SCM $\Mcal=(\Vb, \Ub, \Fcal, \PP_\Ub)$ 
consists of~\citep{pearl2009causality}: 
\begin{enumerate}[label=(\roman*),leftmargin=*,itemsep=-0.1em,topsep=-0.1em]
\item a tuple $\Vb = (V_1, \ldots, V_n)$ of observed, or \textit{endogenous}, variables whose causal relations are modelled;
\item a tuple $\Ub= (U_1, \ldots, U_n)$ of unobserved, or \textit{exogenous}, variables which account for any stochasticity;
\item a tuple $\Fcal = (f_1, \ldots, f_n)$ of deterministic functions, or \textit{mechanisms},
computing each $V_i$ from its \textit{causal parents}, or direct causes, $\PA_i\subseteq\Vb\setminus\{V_i\}$ and the corresponding $U_i$ via the \textit{structural equations}
\vspace{\negspaceeq}
\begin{equation}
\label{eq:structural_equations}
\textstyle
\{V_i := f_i(\PA_i, U_i)\}_{i=1}^n\,;
\vspace{\negspaceeq}
\end{equation}
\item a joint distribution $\PP_\Ub$ over the exogenous $\Ub$.
\end{enumerate}%
Every SCM induces a directed \textit{causal graph}~$\Gcal$ with nodes~$\Vb$ and edges $V_j\rightarrow V_i$ $\forall i, \forall V_j\in \PA_i$ (see~\cref{fig:teaser} for an example).
We make the common assumption that
$\Gcal$ does not contain cycles,\footnote{For a treatment of cyclic SCMs, see~\citet{bongers2021foundations}.} which
ensures that 
$\Mcal$ induces a unique \textit{observational distribution} $\PP_\Vb$ over $\Vb$ (%
see below).
In addition, we assume throughout that all $V_i$ are categorical variables:
\begin{assumption}[Finite domains]
\label{ass:discreteness}
The domains $\Vcal_i$ of all endogenous variables $V_i$ are finite, $\forall i: |\Vcal_i|<\infty$.
\end{assumption}
Whereas for \textit{general} 
SCMs the $f_i$ are arbitrary unknown functions and the domains $\Ucal_i$ of the exogenous $U_i$ unspecified and potentially infinite, \cref{ass:discreteness} permits an \textit{equivalent representation} that makes such SCMs easier to study.
The key observation is that, for categorical $\Vb$, there are only finitely many
functions $\{\ft_{i,k}\}_k$ mapping $\PA_i$ to~$V_i$.
For each value~$u_i$, the function $f_i(\cdot, u_i)$ corresponds to one such \textit{response function}~$\ft_{i,k}$, so $U_i$ acts as a ``random switch'' that induces a distribution on
$\{\ft_{i,k}\}_k$.
We can thus partition the domain $\Ucal_i$ into equivalence classes of values yielding the same $\ft_{i,k}$ and replace $U_i$ with a \textit{categorical response function variable}~$R_i$ defined over these equivalence classes:%
\vspace{\negspaceeq}
\begin{equation}
    \label{eq:response_function_framework}
    \{V_i:=\ft_{i,R_i}(\PA_i)\}_{i=1}^n, \qquad \Rb\sim\PP_\Rb
\vspace{\negspaceeq}
\end{equation}
with $\Rb=(R_1, ..., R_n)$ and each $R_i$ taking values in
\vspace{\negspaceeq}
\begin{equation}
    \label{eq:number_of_response_functions}
    \Rcal_i=
    \{0,...,|\Vcal_i|^{\prod_{V_j\in\PA_i}|\Vcal_j|}-1\}
    \vspace{\negspaceeq}
\end{equation}
\text{if} $\PA_i\neq\varnothing$, and   $\Rcal_i=\{0,...,|\Vcal_i|-1\}$ otherwise.

This re-parametrisation of discrete SCMs is known as the \textit{response function framework}\footnote{also referred to as {\em principal stratification}~\citep{frangakis2002principal} or {\em canonical representation}~\citep{peters2017elements}} 
and we refer to~\citet{balke1994counterfactual} for further details.
Its main benefit is that the $\ft_{i,k}$ are easily enumerated, so that the categorical SCM~\eqref{eq:response_function_framework}
is \text{entirely characterised} by the unknown distribution $\PP_\Rb$.

Using the shorthand $\PP(\xb)$ for $\PP_\Xb(\Xb=\xb)$, the \textit{observational distribution} $\PP_\Vb$ induced by~\eqref{eq:response_function_framework} is given 
by:
\vspace{\negspaceeq}
\begin{equation}
\textstyle
    \label{eq:observational_distribution}
    \PP(\vb)=\medop\sum_{\rb}\PP(\rb)\medop\prod_{i=1}^n\II\{v_i=\ft_{i,r_i}(\pa_i)\}
\vspace{\negspaceeq}
\end{equation}
where $\II\{\cdot \}$ denotes the indicator function.
If $\PP_\Vb$
is known from empirical observation, 
\eqref{eq:observational_distribution}~imposes a constraint on the space of allowed SCMs parametrised by different $\PP_\Rb$.
\textit{Interventions} in the form of external manipulations to subsets~$\Vb_\Ical\subseteq\Vb$ of variables correspond to changes to the structural equations~\eqref{eq:structural_equations}: e.g., setting $\Vb_\Ical$ to a constant $\vb_\Ical$ is denoted using Pearl's \textit{do-operator} by $do(\Vb_\Ical:=\vb_\Ical)$, or $do(\vb_\Ical)$ for short. Interventional distributions are given by:
\vspace{\negspaceeq}
\begin{equation}
    \label{eq:interventional_distribution}
    \PP(\vb_{\setminus \Ical}|do(\vb_\Ical))=\medop\sum_{\rb}\PP(\rb)\medop\prod_{i\not\in\Ical}\II\{v_i=\ft_{i,r_i}(\pa_i)\}
    \vspace{\negspaceeq}
\end{equation}
\textit{Counterfactuals} which condition on some observation $\wb$ of a subset of variables $\Wb \subseteq \Vb$ when reasoning about a hypothetical intervention $do(\vb_\Ical)$ are modelled by using the posterior $\PP_{\Rb|\wb}$ computed via~\eqref{eq:observational_distribution} in place of $\PP_\Rb$ in~\eqref{eq:interventional_distribution}.
Note that the condition $\Wb=\wb$ can contradict the assignment $\Vb_\Ical:=\vb_\Ical$, which renders the query counterfactual.

For now, we additionally make the following common assumption (which we will relax again in~\cref{sec:extensions}).
\begin{assumption}[Causal sufficiency]
\label{ass:causal_sufficiency}
The exogenous variables are mutually independent, i.e., $\PP_\Rb$ factorises.
\end{assumption}%
\Cref{ass:causal_sufficiency} means that there is no hidden confounding, i.e., no unobserved variable influences more than one $V_i$. 
It implies the following \textit{Markov factorisation}~\citep{Spirtes1993}:
\vspace{\negspaceeq}
\begin{equation}
\textstyle
    \label{eq:Markov_factorisation}
    \PP(\vb)=\medop\prod_{i=1}^n \PP(v_i|\pa_i)
    \, ,
\vspace{\negspaceeq}
\end{equation}
where each \textit{causal Markov kernel} $\PP(v_i|\pa_i)$ is given by
\vspace{\negspaceeq}
\begin{equation}
\label{eq:causal_Markov_kernel}
\textstyle
    \PP(v_i|\pa_i)=\medop\sum_{r_i\in\Rcal_i}\PP(r_i) \, \II\{v_i=\ft_{i,r_i}(\pa_i)\}
    \,.
\vspace{\negspaceeq}
\end{equation}
\Cref{ass:causal_sufficiency} has two important consequences: first, it suffices to consider the marginals of each $R_i$ separately (rather than model their joint distribution $\PP_\Rb$); second, interventional queries become identifiable from observational data via the g-formula~\citep{robins1986new}, a.k.a.\ \textit{truncated factorisation}
\vspace{\negspaceeq}
\begin{equation}
\textstyle
    \label{eq:truncated_factorisation}
    \PP(\vb_{\setminus \Ical}|do(\vb_\Ical))=\delta(\vb_\Ical) \medop\prod_{i\not\in\Ical}\PP(v_i|\pa_i)
    \,.
    \vspace{\negspaceeq}
\end{equation}
Under~\cref{ass:causal_sufficiency}, the boundary between interventional (rung~2) and observational (rung~1) quantities thus disappears once the causal graph is known.  However, there is typically still a whole family of SCMs consistent with the available rung~1/2 information that imply different counterfactuals (rung~3), see, e.g., ~\citet[][\S~3.4]{peters2017elements} for an explicit description of this ambiguity.
Next, we illustrate this point for Boolean SCMs which will be the main objects of study. 

\subsection{Causally-Sufficient Cause-Effect Models}
\label{sec:Boolean_cause_effect}
Consider a bivariate, Boolean SCM $\MA$ over $X\to Z$. Using response functions, this can be written \textit{w.l.o.g.}\ as
\vspace{\negspaceeq}
\begin{equation}
\label{eq:MA}
X:=R_X, \qquad 
Z:=f_{R_Z}(X), 
\vspace{\negspaceeq}
\end{equation}
where $R_Z$ indexes the four distinct functions $f_k$ from $\{0,1\}$ to $\{0,1\}$: 
the two constant functions
$f_0\equiv 0$, and $f_1\equiv1$, as well as $f_2(X)=X$ (``ID''), and $f_3(X)=1-X$~(``NOT'').

Here, 
$\PP_{R_X}$ coincides with the (observed) marginal $\PP_X$, and
we assume that $X$ is not constant, $0<\PP(X=1)<1$. 
Under~\cref{ass:causal_sufficiency}, the SCM $\MA$ from~\eqref{eq:MA} is thus characterised entirely by the distribution $\PP_{R_Z}$ over the four~$f_k$. We represent this as a probability vector $\ab\in\Delta^3$, where $\Delta^{K-1}=\{\ab \in \mathbb{R}^K \, | \, a_k \geq 0, \sum_{k=0}^{K-1} a_k = 1\}$ denotes the probability simplex over $K$ points.
Due to the constraints imposed on~$\ab$ by the observed~$\PP_{Z|X}$ via~\eqref{eq:causal_Markov_kernel}, we can write it in terms of a \textit{single free parameter}~$\lamA\in[\lamA^\text{min},\lamA^\text{max}]$ as:
\vspace{\negspaceeq}
\begin{equation}
\label{eq:lam_A}
\textstyle
\ab(\lamA)
=
\begin{pmatrix}
0\\
1
-
p_{00}
-
p_{01}\\
p_{00}\\
p_{01}
\end{pmatrix}
+\lamA
\begin{pmatrix}
1\\
1\\
-1\\
-1
\end{pmatrix}
\vspace{\negspaceeq}
\end{equation}
with $p_{ij}=\PP(Z=i|X=j)$, $\lamA^\text{min}=\max\{0, p_{00}+p_{01}-1\}$, and $\lamA^\text{max}=\min\{p_{00},p_{01}\}$, see%
~\cref{app:single_param_scm_s}
for details. 

Different choices of~$\lamA\in[\lamA^\text{min},\lamA^\text{max}]$ thus define a \textit{family of SCMs} that are observationally and interventionally equivalent\footnote{i.e., indistinguishable based on all $do$-interventions.%
} but imply \textit{different counterfactual distributions}.
In particular, for any given observation $(x,z)$, the probability that ``$Z$ would have flipped had $X$ been different'' is given by $\gamma(\lamA) :=a_2+a_3=p_{00} + p_{01} - 2\lamA$. 
\looseness-1 For this reason, we call $\gamma$ the \emph{counterfactual influence} of $X$ on $Z$. 
SCMs with larger $\lamA$ thus exhibit a smaller counterfactual influence.%

\section{The Structural Causal Marginal Problem}
\label{sec:counterfactual_marginal_problem}
We now formulate the
\textit{causal marginal problem}, which can be understood as a causal version of the (statistical) marginal problem~\citep{vorob1962consistent,Kellerer1964}.

\begin{tcolorbox}[boxsep=-0.25em]
\textbf{Causal Marginal Problem:}
\textit{Can marginal causal models 
over subsets of variables with
known causal graph be consistently merged? What constraints on marginal and joint causal models does this imply?}
\end{tcolorbox}

We study this problem within the SCM framework, i.e., the \textit{structural causal marginal problem}.
To build intuition
and gain a better understanding of the fundamental concepts,
we first analyse
the causally-sufficient, Boolean setting from~\cref{sec:Boolean_cause_effect}: we assume that in addition to $\PP_{XZ}$, we observe $\PP_{YZ}$ from another dataset where $Y$ is a second independent Boolean cause of $Z$, as illustrated in~\cref{fig:teaser}.\footnote{\Cref{ass:causal_sufficiency} implies $X\indep Y$, for otherwise $Y$ (resp.\ $X$), which is unobserved in $\MA$ (resp.\ $\MB$), would be a hidden confounder. This is, in principle, falsifiable through observation of $\PP_{XY}$.}
Crucially, we do not have joint observations of all three variables, i.e., $\PP_{XYZ}$ is unknown.
While this case might appear rather simple, it already bears a number of nontrivial implications for counterfactual inference.
We defer a more general definition and a discussion of extensions to~\cref{sec:extensions}.

We denote the second marginal SCM over $Y\rightarrow Z$ by~$\MB$,%
\vspace{\negspaceeq}
\begin{equation}
\label{eq:MB}
Y:=Q_Y, \qquad 
Z:=f_{Q_Z}(Y), 
\vspace{\negspaceeq}
\end{equation}
using the same response functions $f_k$ as in~\eqref{eq:MA} for $\MA$,
and parametrise the family of SCMs consistent with
the observed $\PP_{YZ}$ with a probability vector~$\bb\in\Delta^3$ with a single free parameter~$\lamB\in[\lamB^\text{min},\lamB^\text{max}]$, analogously to~\eqref{eq:lam_A}.

The space of marginal SCMs $(\MA,\MB)$ parametrised by $(\lamA,\lamB)$ that are \textit{separately} consistent with $\PP_{XZ}$ and $\PP_{YZ}$ (i.e., prior to considerations about consistently merging them into a joint model) is illustrated in~\cref{fig:overview} as the red rectangle.
We will show that: (i) enforcing 
that the two marginal SCMs can be merged into a joint SCM~(\cref{sec:projections}) reduces the space of admissible $(\MA,\MB)$ (blue \& green areas in~\cref{fig:overview}; \cref{subsec:solutions,sec:experiments}); (ii) knowing one of the marginal SCMs exactly (e.g., from prior knowledge or particular observations%
) further restricts the choices for the other marginal (horizontal green line in~\cref{fig:overview}; \cref{sec:examples}); and (iii) some marginal models are inherently easier to falsify than others~(\cref{sec:theory,sec:experiments}).

\subsection{Consistency Between Marginal and Joint SCMs}
\label{sec:projections}
We now define the joint model and provide a systematic way of linking its representation to those of the marginal models. 
We write the joint SCM~$\MC$ over $\{X,Y\}\rightarrow Z$ as
\vspace{\negspaceeq}
\begin{equation}
\label{eq:MC}
X:=R_X, \qquad
Y:=Q_Y, \qquad 
Z:=h_{S}(X,Y), 
\vspace{\negspaceeq}
\end{equation}
where $S$ indexes the $16$ response functions $h_0, \ldots, h_{15}$ from $\{0,1\}^2$ to $\{0,1\}$ (listed in~\cref{tab:15functions} in~\Cref{app:joint_model_parametrisation}).
We denote the distribution $\PP_{S}$ over the $h_k$ by a probability vector~$\cb\in\Delta^{15}$.
Note that unlike for the marginal models, we do not a priori have additional constraints reducing the number of free parameters of $\cb$ since $\PP_{XYZ}$ and thus $\PP_{Z|XY}$ are unknown.

\newcommand{\maxA}{7}
\newcommand{\maxB}{5}
\newcommand{\lbA}{0.075*\maxA}
\newcommand{\ubA}{0.925*\maxA}
\newcommand{\lbB}{0.075*\maxB}
\newcommand{\ubB}{0.9*\maxB}
\newcommand{\lamAmin}{0.2*\maxA}
\newcommand{\lamAmax}{0.75*\maxA}
\newcommand{\lamBmin}{0.25*\maxB}
\newcommand{\lamBmax}{0.75*\maxB}
\newcommand{\eps}{0.05} %
\newcommand{\polyy}{0.6*\maxB}
\newcommand{\polyx}{0.35*\maxA}
\newcommand{\polyyy}{0.5*\maxB}
\newcommand{\polyxx}{0.4*\maxA}
\newcommand{\lamBstar}{0.375*\maxB}
\newcommand{\xgreen}{0.5*\lamAmin+0.5*\lamAmax}
\newcommand{\xyellow}{0.925*\lamAmax}
\newcommand{\xred}{1.1*\lamAmax}

\begin{figure}[t!]
    \centering
    \begin{tikzpicture}
    \centering

    \draw[-] (0,0) -- (0,0) node[anchor= north east] {$0$};

    \draw[->, thick] (0,0) -- (\maxA,0) node[anchor= west] {$\lamA$};

    \draw[->, thick] (0,0) -- (0,\maxB) node[anchor= south] {$\lamB$};

    \fill[color=cb-burgundy!20!white] (\lbA,\lbB) rectangle (\ubA,\ubB);
    \draw[-,ultra thin] (\lbA,\lbB) rectangle (\ubA,\ubB);
    \pattern[pattern color=cb-burgundy!50!white, pattern=vertical lines] (\lbA,\lbB) rectangle (\ubA,\ubB);

    \fill[color=cb-blue!20!white] (\lamAmin,\lamBmin) rectangle (\lamAmax,\lamBmax);
    \draw[-,ultra thin] (\lamAmin,\lamBmin) rectangle (\lamAmax,\lamBmax);

    \coordinate (P1) at (\lamAmin,\polyy);
    \coordinate (P2) at (\polyx,\lamBmax);
    \coordinate (P3) at (\lamAmax, \polyyy);
    \coordinate (P4) at (\polyxx, \lamBmin);

    \draw[fill=cb-green-lime!20!white, ultra thin] (P1) -- (P2) -- (P3) -- (P4) -- cycle;
    \pattern[pattern color=cb-green-sea!50!white, pattern=dots] (P1) -- (P2) -- (P3) -- (P4) -- cycle;

    \draw[-, dashed, thick, color=cb-green-sea, name path=lamBstar] (0,\lamBstar) -- (\maxA,\lamBstar);

    \draw[name path=diag_min] (P1) -- (P4);
    \draw[name path=diag_max] (P3) -- (P4);
    \path[name intersections={of = lamBstar and diag_min}];
    \coordinate (lam_star_min)  at (intersection-1);
    \path[name intersections={of = lamBstar and diag_max}];
    \coordinate (lam_star_max)  at (intersection-1);
    \draw[-, ultra thick, color=cb-green-sea] (lam_star_min) -- (lam_star_max);
    \draw[-] let \p1 = (lam_star_min) in (\x1,\y1-3*\eps) -- (\x1,\y1+3*\eps);
    \draw[-, ultra thick, color=cb-green-sea] let \p1 = (lam_star_min) in (\x1,\lamBstar-2*\eps) -- (\x1,\lamBstar+2*\eps);
    \draw[-, ultra thick, color=cb-green-sea] let \p1 = (lam_star_max) in (\x1,\lamBstar-2*\eps) -- (\x1,\lamBstar+2*\eps);

    \draw[-] (\lbA,-\eps) -- (\lbA,\eps) node[anchor= north] {$\lamA^\text{min}$};
    \draw[-] (\ubA,-\eps) -- (\ubA,\eps) node[anchor= north] {$\lamA^\text{max}$};
    \draw[-] (-\eps,\lbB) -- (\eps,\lbB) node[anchor= east] {$\lamB^\text{min}$};
    \draw[-] (-\eps,\ubB) -- (\eps,\ubB) node[anchor= east] {$\lamB^\text{max}$};
    \draw[-] (\lamAmin,-\eps) -- (\lamAmin,\eps) node[anchor= north, yshift=-0.1em] {$\LBA$};
    \draw[-] (\lamAmax,-\eps) -- (\lamAmax,\eps) node[anchor= north, yshift=-0.1em] {$\UBA$};
    \draw[-] (-\eps,\lamBmin) -- (\eps,\lamBmin) node[anchor= east] {$\LBB$};
    \draw[-] (-\eps,\lamBmax) -- (\eps,\lamBmax) node[anchor= east] {$\UBB$};
    \draw[-] (-\eps,\lamBstar) -- (\eps,\lamBstar) node[anchor= east] {$\lamB^\star$};
    \draw[-] let \p1 = (lam_star_min) in (\x1,-\eps) -- (\x1,\eps) node[anchor= north] {$\LBA^\star$};
    \draw[-] let \p1 = (lam_star_max) in (\x1,-\eps) -- (\x1,\eps) node[anchor= north] {$\UBA^\star$};
    \draw[-] (3.5,3) -- (3.5,3) node[anchor= north east] {\LARGE $\Lambda_\Ccal$};
    
    \draw[-] (6.25,4) -- (6.25,4) node[anchor= north east] {\LARGE $\Lambda_0$};
    \node[star, fill, color=cb-blue, scale=0.5] (Myellow) at (\xyellow,\lamBstar) {};
    \end{tikzpicture}
    \caption{%
    \looseness-1
    \textbf{2D Schematic of the Structural Causal Marginal Problem.} 
    For the causal graph from~\cref{fig:teaser} and model class from~\cref{sec:Boolean_cause_effect}, given $\PP_{XZ}$ and $\PP_{YZ}$, the two marginal SCMs $\MA$ and $\MB$ over $X\to Z$ and $Y\to Z$ are each parametrised by a single free parameter $\lamA\in[\lamA^\text{min},\lamA^\text{max}]$ (x-axis) and $\lamB\in[\lamB^\text{min},\lamB^\text{max}]$ (y-axis), respectively.
    The outer dashed red  area~$\Lambda_0$ corresponds to combinations of counterfactual marginal models $(\lamA,\lamB)$ that are falsified in that they cannot be counterfactually consistent~(\cref{def:counterfactual_consistency}); 
    the inner dotted green polytope~$\Lambda_\Ccal$ corresponds to $(\lamA,\lamB)$ that are counterfactually consistent;
    and the solid blue area, defined as the surrounding rectangle of the latter, corresponds to $(\lamA,\lamB)$ that are not counterfactually consistent but cannot be falsified without additional assumptions or constraints.
    Enforcing consistency with the other marginal (interventional) model implies $\lamA\in[\LBA,\UBA]$ and $\lamB\in[\LBB,\UBB]$, but without additional information  about the other marginal this range cannot be reduced further. 
    For a given~$\MB$ corresponding to~$\lamB^\star$ (dashed horizontal green line), on the other hand, the interval of consistent $\lamA$ shrinks further to $[\LBA^\star,\UBA^\star]$ (solid green line) so that the $\lamA$ corresponding to the blue star marker can be ruled out. 
    }
    \label{fig:overview}
\end{figure}

To relate the joint~\eqref{eq:MC} and marginal SCMs~\eqref{eq:MA} and~\eqref{eq:MB},
a key observation is that for any fixed value $x$ of $X$ (resp.\ $y$ of $Y$),
each two-variable function $h_k(X,Y)$   implicitly defines a single-variable function $f_j(Y)$ over the remaining variable $Y$ (resp.\ $f_{j'}(X)$ over $X$).
Formally, we define the following projection operators for $x,y\in\{0,1\}$:
\vspace{\negspaceeq}
\begin{equation}
\label{eq:projection_operators}
\begin{aligned}
    \Pcal_x^X&: h_k
    \mapsto h_k(x,Y)=f_j(Y) \quad \text{for some} \quad j,
    \\
    \Pcal_y^Y&: h_k
    \mapsto h_k(X,y)=f_{j'}(X) \quad \text{for some} \quad j'.
\end{aligned}
\vspace{\negspaceeq}
\end{equation}
For example, we defined $h_0(X,Y)\equiv 0$
(see~\cref{tab:15functions} in~\cref{app:joint_model_parametrisation}), 
so $\Pcal^X_0(h_0)=f_0(Y)\equiv 0$;
and, 
similarly, for $h_7(X,Y) = \neg(X\wedge Y)$ we have that $\Pcal^Y_1(h_7)=f_3(X)$ since $h_7(X, 1)=1-X$, i.e., the NOT function~$f_3$.

Together with the marginal distributions of $X$ and $Y$ (obtained by marginalisation of $Z$ in $\PP_{XZ}$ and $\PP_{YZ}$), the distribution over the $h_k$ in $\MC$ parametrised by $\cb\in\Delta^{15}$ thus induces distributions over the $f_j$ in $\MA$ and $\MB$ via~\eqref{eq:projection_operators}. 
The latter are parametrised by $\ab(\lamA)$ and $\bb(\lamB)$ (see~\eqref{eq:lam_A}), and enforcing that they match the corresponding distributions induced by $\MC$ yields the following \textit{linear} constraints:%
\vspace{\negspaceeq}
\begin{equation}
\label{eq:constraints_scalar}
\begin{aligned}
\textstyle
    a_j(\lamA)&=
    \medop \sum_{y=0}^1\PP_Y(y)\medop \sum_{k=0}^{15}\II\{\Pcal^Y_y(h_k)=f_j(X)\} \, c_k\,,
    \\
    b_j(\lamB)&=
    \medop\sum_{x=0}^1\PP_X(x)\medop\sum_{k=0}^{15}\II\{\Pcal^X_x(h_k)=f_j(Y)\} \, c_k\,,
\end{aligned}
\vspace{\negspaceeq}
\end{equation}
for $j=0,1,2,3$.
Writing~\eqref{eq:constraints_scalar} in matrix form, we obtain:
\vspace{\negspaceeq}
\begin{equation}
\label{eq:constraints_matrix}
    \begin{aligned}
    \ab(\lamA)= \Ab
    \cb, 
    \qquad \qquad 
    \bb(\lamB)= \Bb
    \cb\,,
    \end{aligned}
\vspace{\negspaceeq}
\end{equation}
where {$\Ab,\Bb\in\RR^{4\times 16}$} are constant matrices whose entries are given in terms of $\PP_Y$ and $\PP_X$, respectively.\footnote{Specifically,
$(\Ab)_{jk}=\sum_{y=0}^1\PP_Y(y)\II\{\Pcal^Y_y(h_k)=f_j(X)\}$, and 
$(\Bb)_{jk}=\sum_{x=0}^1\PP_X(x)\II\{\Pcal^X_x(h_k)=f_j(Y)\}$.
}

In general, \eqref{eq:constraints_matrix} does not uniquely determine a joint SCM in terms of the marginal ones as it involves at most eight independent constraints. Nor does there always exist a joint model (parametrised by $\cb$) that satisfies~\eqref{eq:constraints_matrix}: take, e.g., any combination of $\PP_{XZ}$ and $\PP_{YZ}$ for which already the \textit{statistical} marginal problem does not have a solution.

To discuss solutions to the structural causal marginal problem, we introduce the following notion of consistency.

\begin{definition}[Counterfactual consistency]
\label{def:counterfactual_consistency}
An SCM $\Mcal$ over observed variables $\Vb$ is \textit{counterfactually consistent} with a (marginal) SCM $\Mcal_1$ over a subset $\Wb_1\subseteq\Vb$ if all counterfactual distributions of $\Wb_1$ in $\Mcal_1$ coincide with those implied by $\Mcal$ via marginalisation of $\Vb\setminus \Wb_1$, \citep[see][Defn.~5.3 for  marginalisation of SCMs]{bongers2021foundations}.
Two SCMs $\Mcal_1,\Mcal_2$ over subsets $\Wb_1,\Wb_2\subseteq\Vb$ are counterfactually consistent if there is a joint SCM $\Mcal$ over $\Vb$ which is counterfactually consistent with both $\Mcal_1$ and~$\Mcal_2$.
\end{definition}%
\Cref{def:counterfactual_consistency} can be understood as a generalisation of
counterfactual \textit{equivalence}~\citep[see, e.g.,][Defn.~6.47]{peters2017elements} which also involves equality of counterfactual distributions, but applies to different SCMs over the \textit{same} set of variables.

The counterfactual distributions implied by an SCM are fully determined by the structural equations and noise distribution (as parametrised by $\lamA$, $\lamB$, and $\cb$ here).
In our case, a marginal SCM $\MA$ (or $\MB$) is thus counterfactually consistent with a joint SCM $\MC$ if the corresponding constraint in~\eqref{eq:constraints_matrix} holds. 
The two marginal SCMs $\MA$ and $\MB$ are counterfactually consistent if both constraints in~\eqref{eq:constraints_matrix} hold simultaneously for some $\cb$.
In this case, we say that $\cb$ (or $\Mcal$)
is a \textit{solution} to the structural causal marginal problem.

\subsection{Determining the Space of Solutions}
\label{subsec:solutions}
As discussed, \eqref{eq:constraints_matrix}
and the simplex constraints $\cb\in\Delta^{15}$, and $(\lamA, \lamB)\in[\lamA^\text{min},\lamA^\text{max}]\times[\lamB^\text{min},\lamB^\text{max}]=:\Lambda_0$
define the solution space for the structural causal marginal problem.
\looseness-1 Specifically, they imply a set of \textit{linear} equality and inequality constraints that, if satisfiable, yield a \textit{convex polytope}~$\Ccal$ %
as the feasible set for~$\cb$~\cite{boyd2004convex}:%
\vspace{\negspaceeq}
\begin{equation}\label{eq:C_polytope}
    \mathcal{C}:= \{\cb \in \Delta^{15} \mid 
    \exists (\lamA, \lamB) \in \Lambda_0 \,\,\,
    \text{s.t.} \,\,\, \eqref{eq:constraints_matrix} \,\,\, \text{holds}
    \}
\vspace{\negspaceeq}
\end{equation}
see~\cref{app:reformulation_linear_program} for details.
By \eqref{eq:lam_A} and \eqref{eq:constraints_matrix}, we have that $\lamA = [\Ab\cb]_0$ and $\lamB = [\Bb\cb]_0$, so 
the set of jointly feasible  $(\lamA,\lamB)$ is given by $\Lambda_\Ccal := \{\left([\Ab\cb]_0, \, [\Bb\cb]_0 \right)^\top \mid \cb \in \mathcal{C}\}$.
$\Lambda_\Ccal$ is illustrated as the dotted green region in~\cref{fig:overview}.

We could now minimise and maximise some (linear) causal query $\Qcal(\cb)$ over $\cb\in\Ccal$,
to obtain bounds on counterfactuals of interest, e.g., the ETT for Alice mentioned in~\cref{sec:intro}.
Since $\Ccal$ is convex, this results in a linear program which can be solved easily and with global optimality guarantees~\cite{dantzig1963linear,karmarkar1984new}.
Such an approach is closely related to partial identification~(cf.~\cref{sec:intro}).
Here, we focus instead on how the space of marginal and joint models is reduced when additional marginals are observed.

Does enforcing counterfactual consistency meaningfully restrict the space of admissible marginal SCMs?
To check this, we can compare, e.g., the interval $[\lamA^\text{min},\lamA^\text{max}]$ of allowed~$\lamA$ \textit{prior} to enforcing~\eqref{eq:constraints_matrix} 
with the lower and upper bounds $[\LBA,\UBA]$ defined as 
$
\minmax_{
\Lambda_\Ccal} \lamA$, and similarly for $[\LBB, \UBB]$. 
The 
region $[\LBA,\UBA]\times[\LBB,\UBB]$ 
is illustrated as the solid blue area in~\cref{fig:overview}. 
\looseness-1
By definition, it is the rectangle delimiting the projection $\Lambda_\Ccal$ of  the polytope $\Ccal$ of feasible
solutions in the $(\lamA,\lamB)$-plane. 
If the blue and dashed red rectangles coincide, neither of the marginal SCMs is further restricted by enforcing consistency.
Otherwise, marginals that fall outside the blue area are falsified in that they cannot be counterfactually consistent.

We highlight a subtle point regarding the blue area in~\cref{fig:overview},  counterfactual consistency, and falsifiability: 
If $(\lamA,\lamB)$ lies within the blue region but outside $\Lambda_\Ccal$ (e.g., the blue star marker in~\cref{fig:overview}), the corresponding marginal SCMs $\MA$ and $\MB$ are not counterfactually consistent. However, neither of them is therefore falsified%
; it is only their combination that can be ruled out.
Since we generally know neither of the marginal SCMs exactly (assuming we only observe $\PP_{XZ}$ and $\PP_{YZ}$), 
for any $\lamA\in[\LBA,\UBA]$, by definition, there is a $\lamB'$ such that $(\lamA,\lamB')$ are counterfactually consistent.
Hence, $\lamA$ cannot be ruled out without additional knowledge about~$\lamB$.
If, on the other hand, we know that $\lamB=\lamB^\star$ (illustrated as the horizontal dashed green line in~\cref{fig:overview}), the red rectangle degenerates to the interval $[\lamA^\text{min}, \lamA^\text{max}]\times\{\lamB^\star\}$, and the blue and green regions coincide and collapse to the sub-interval $[\LBA^\star,\UBA^\star]\times\{\lamB^\star\}$ defined as $\minmax_{(\lamA,\lamB^\star)\in\Lambda_\Ccal} \lamA$,  shown as the solid green interval in~\cref{fig:overview}. 
Next, we illustrate this with an example.

\newcommand\height{4.5}
\newcommand\leftplace{10}
\newcommand\heightcapt{5}
\begin{figure*}[t]
    \begin{subfigure}{0.29\textwidth}
        \centering

        \begin{overpic}[height=\height cm]{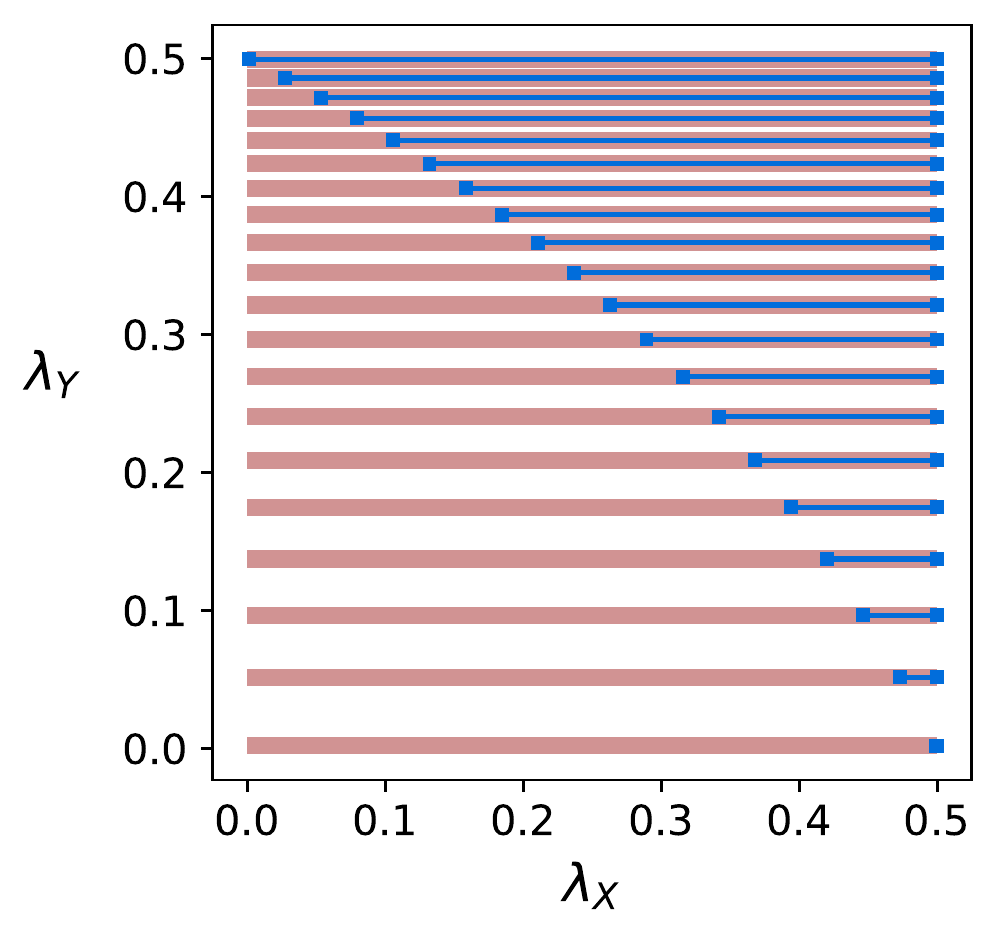}
         \put (\leftplace, \heightcapt) {\textbf{\small(a)}}
\end{overpic}
    \end{subfigure}
    \begin{subfigure}{0.35\textwidth}
        \centering
        \begin{overpic}
        [height=\height cm]{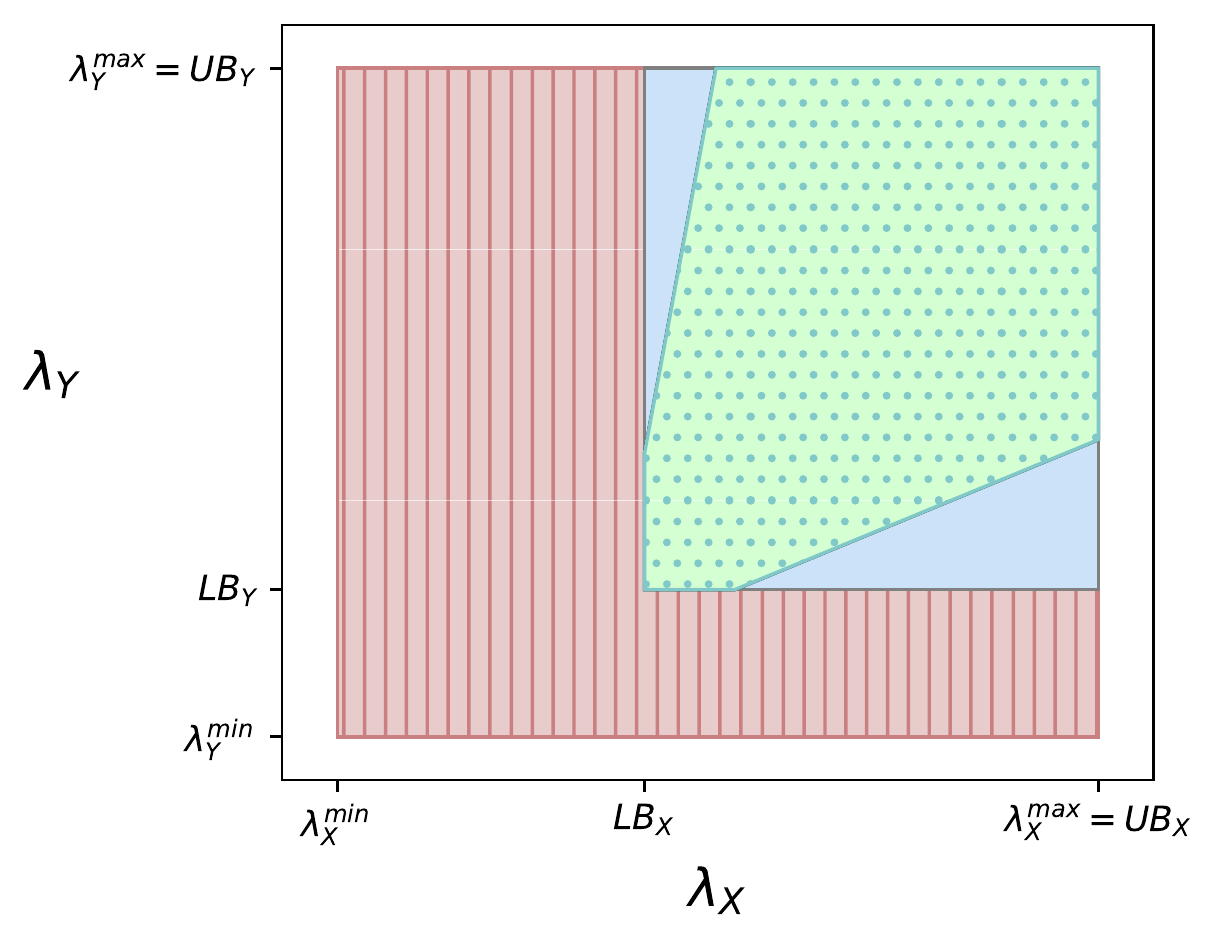}
        \put (\leftplace, \heightcapt) {\textbf{\small(b)}}
\end{overpic}
    \end{subfigure}
    \begin{subfigure}{0.35\textwidth}
        \centering
        \begin{overpic}[height=\height cm]{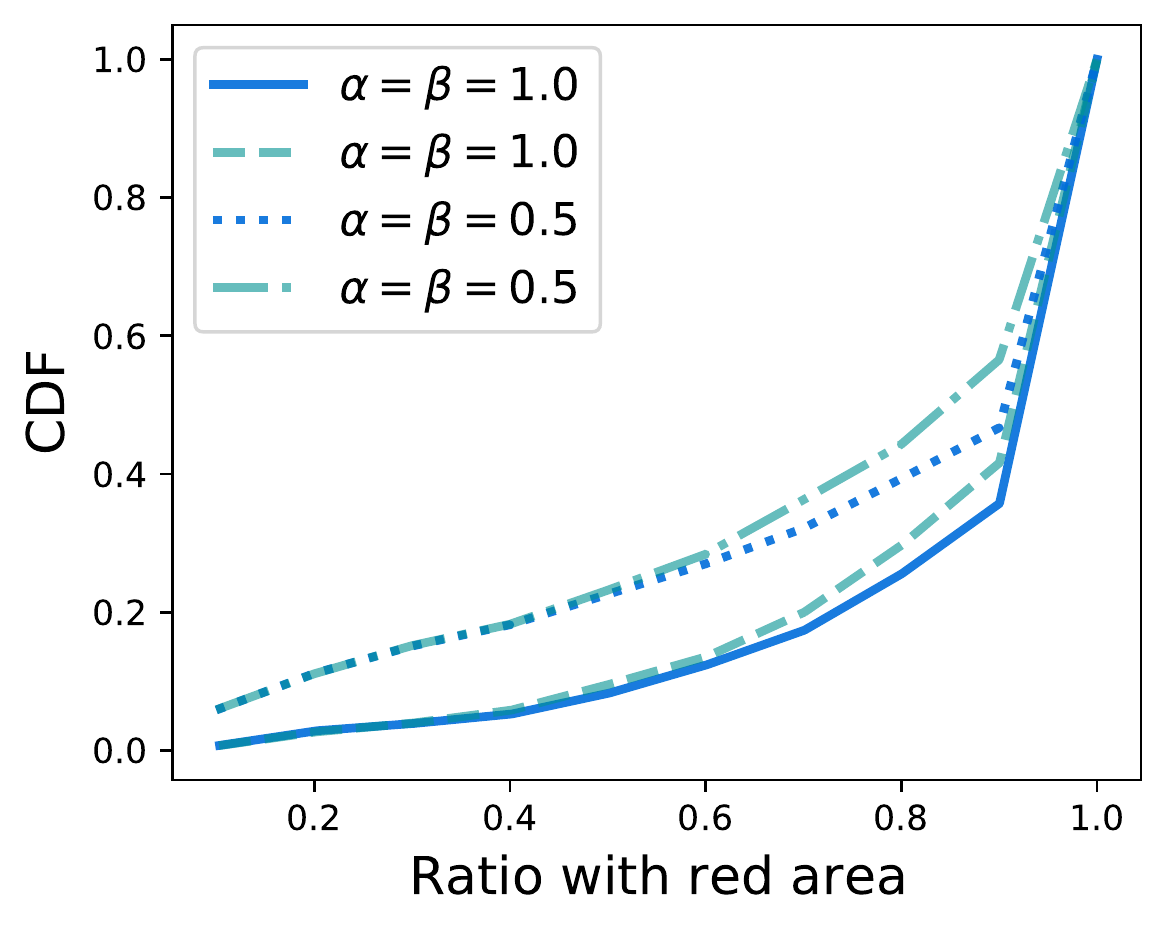}
        \put (\leftplace, \heightcapt) {\textbf{\small(c)}}
    \end{overpic}
    \end{subfigure}
    \caption{%
    \textbf{(a)}~For the example from~\cref{sec:examples}, both the unique allowed value of $\lamB$ and the range of consistent $X \to Z$ models~$\MA$ (thin blue lines) change as $\theta=\PP(Y=1)$ is varied. 
    \textbf{(b)}~An instance %
    of the %
    structural causal marginal problem that entails constraints for both $\lamA$ and $\lamB$.
    Note that $(\lamA^{\text{max}}, \lamB^{\text{max}}) \in \Lambda_\Ccal$, as implied by~\cref{prop:lambda_max}. %
    \looseness-1 \textbf{(c)} In solid/dotted blue (resp.~dashed green), 
    CDFs of the observed ratios between the blue (resp.~green)  and red area %
    for different
    Beta priors 
    over $\PP_{Z|XY}$: often, counterfactual consistency induces meaningful constraints. 
    }
    \label{fig:experiments}
\end{figure*}

\subsection{Worked-Out Example}
\label{sec:examples}

Suppose that the observed marginal distributions $\PP_{XZ}$ and $\PP_{YZ}$ are such that they satisfy the following:
(i)~$X\indep Z$,
(ii)~$\PP(Y=0, Z=1)=0$,
(iii)~$\PP(Z=1)=0.5$, and
(iv)~$0<\PP(X=0),\PP(Y=0)<1$.
Crucially, these assumptions only involve empirical quantities and do not require joint observations of $(X, Y, Z)$.
We focus on the main points here and refer to~\cref{app:examples} for detailed derivations.

First, we consider the $Y\to Z$ marginal SCM $\MB$ in~\eqref{eq:MB}. 
Assumption (ii) implies zero probability for the constant one ($f_1$) and NOT ($f_3$) functions.
Together with (iii), it turns out that this \textit{uniquely} determines $\MB$: we must have $\lamB=\lamB^\star:=\frac{2\theta-1}{2\theta}$ where $\theta=\PP(Y=1)$, and the response function distribution is  given by $\bb=(\frac{2\theta-1}{2\theta}, 0, \frac{1}{2\theta}, 0)^\top$.\footnote{Note that (ii) and (iii) together imply that $\theta\geq 0.5$ since $\PP(Z=1)=\PP(Y=1, Z=1)=\theta \, \PP(Z=1|Y=1)=0.5$}
This can also be written more compactly as an AND model:%
\vspace{\negspaceeq}
\begin{equation}
\label{eq:example_and_model}
\textstyle 
    Z:=Y\land N_Z, 
    \qquad 
    N_Z\sim\text{Bernoulli}(\frac{1}{2\theta}).
\vspace{\negspaceeq}
\end{equation}

Next, we consider the $X\to Z$ marginal SCM $\MA$ in~\eqref{eq:MA}.
Intuitively,  assumption (i) rules out SCMs that do not give equal weight to the constant zero ($f_0$) and one ($f_1$) functions, as well as to the ID ($f_2$) and NOT ($f_3$) functions, for otherwise $X$ and $Z$ could not be statistically independent.
Substituting (i) and (iii) into~\eqref{eq:lam_A}, we indeed find the family of response function distributions implied by~$\PP_{XZ}$ to be  $\ab(\lamA)=(\lamA,\lamA,0.5-\lamA, 0.5-\lamA)^\top$ with 
$0\leq\lamA\leq 0.5$.
For example, for $\lamA=0.5$ this yields%
\vspace{\negspaceeq}
\begin{equation}
    \label{eq:example_indepedent_model}
    Z:=M_Z,
    \qquad
    M_Z\sim\text{Bernoulli}(0.5),
\vspace{\negspaceeq}
\end{equation}
whereas for $\lamA=0$ we obtain
\vspace{\negspaceeq}
\begin{equation}
    \label{eq:example_xor_model}
    Z:=X\oplus M_Z,
    \qquad
    M_Z\sim\text{Bernoulli}(0.5).
\vspace{\negspaceeq}
\end{equation}
As discussed,
\eqref{eq:example_indepedent_model} and~\eqref{eq:example_xor_model}
are interventionally equivalent---$Z$ is an unbiased coin toss regardless of~$X$%
---but entail different counterfactuals:
given some $(x,z)$, the statement ``$Z$ would have been different, had $X$ been $x' \neq x$'' would be true only for the XOR model~\eqref{eq:example_xor_model} but false for~\eqref{eq:example_indepedent_model}. This reflects that for \eqref{eq:example_xor_model} the counterfactual influence is $\gamma=1$, while for \eqref{eq:example_indepedent_model} it is $\gamma=0$.
We also note that~\eqref{eq:example_xor_model} violates faithfulness~\cite{Spirtes1993}.\footnote{Our point could, in principle, also be made for more generic causal models, but the math is less simple then.}

Next, we analyse whether and how the problem is further constrained by enforcing counterfactual consistency.
Recall that in $\MB$ we have $b_1=b_3=0$. 
Together with assumption (iv), the second constraint in~\eqref{eq:constraints_scalar} for $j=1,3$ then implies that all but four of the $c_k$ are zero.
The first constraint in~\eqref{eq:constraints_matrix} then yields a system of four linear equations relating the non-zero components $c_0, c_2, c_8, c_{10}$ of $\cb$ to~$\ab(\lamA)$.
By solving for $c_0$ and enforcing positivity, $c_0\geq0$, we finally obtain the consistency constraint: $1-\theta\leq\lamA$.
In summary, if we \textit{only} observe $\PP_{XZ}$, any $\lamA\in [0, 0.5]$ is allowed; if we \textit{additionally} know $\PP_{YZ}$ and enforce counterfactual consistency, this interval shrinks to $\lamA\in[1-\theta, 0.5]$.
The space of counterfactually consistent $\MA$ can thus be arbitrarily small, depending on $0.5\leq \theta=\PP(Y=1)<1$.
This is illustrated in~\cref{fig:experiments}~\textbf{(a)} for different values of $\theta$, see~\cref{sec:experiments} for details. 
In particular, we note that
the (unfaithful) XOR model~\eqref{eq:example_xor_model} is falsified in that it can never be counterfactually consistent with $\MB$ from~\eqref{eq:example_and_model}.
Moreover, in the extreme case that~$\theta=0.5$, the interval collapses to a point and the only admissible $\MA$ is~\eqref{eq:example_indepedent_model} where $X$ has no counterfactual influence on $Z$. 
This seems intuitive since $\MB$ puts all weight on the ID function ($Z:=Y$) for $\theta=0.5$, i.e., $Y$ fully determines $Z$ in that case.

\subsection{Some Marginal SCMs Cannot Be Falsified}
\label{sec:theory}
In the previous example, enforcing consistency with the interventional $Y\to Z$ marginal only affected the \textit{lower} bound on~$\lamA$. In fact, it can be shown that this holds more generally 
for both $\lamA$ and $\lamB$ (proof in~\cref{app:lambda_max}):
\begin{proposition}\label{prop:lambda_max}
Consider the structural causal marginal problem  described in~\cref{sec:projections,subsec:solutions}, with %
$X \rightarrow Z \leftarrow Y$, causal sufficiency, and Boolean~$X, Y, Z$. If a solution exists (i.e., $\Ccal$~is non-empty),
we  have $(\lamA^\text{max}, \lamB^\text{max})^\top \in \Lambda_{\mathcal{C}}$.
\end{proposition}
In particular, this implies $\UBA =\nolinebreak \lamA^\text{max}$ and $\UBB =\nolinebreak \lamB^\text{max}$, as illustrated in~\cref{fig:experiments}~\textbf{(b)}.%
\footnote{\cref{fig:overview} should thus be understood as a conceptual visualisation rather than an exact representation; \cref{fig:experiments}~\textbf{(b)} is a refinement.} %
As a result, the structural causal marginal problem %
cannot falsify models $\MA$ or $\MB$ that assign the maximally allowed weight to the constant functions $Z:=0$ and $Z:=1$. Conversely, models corresponding to small values of $\lamA,\lamB$ can sometimes be falsified: note that these are the models where the cause $X$ (or $Y$) has a stronger counterfactual influence on $Z$, as defined in~\cref{sec:Boolean_cause_effect}. We elaborate on the significance of this result in~\cref{sec:discussion}.

\section{Experiments}
\label{sec:experiments}
In~\cref{fig:experiments}~\textbf{(a)} we visualise
the worked-out example from~\cref{sec:examples}. %
Recall that there the interventional $Y\to Z$ model  uniquely determines~$\MB$, and $\Lambda_0$ is therefore a segment with~$\lamA \in [0, 0.5]$ and $y$-coordinate fixed by~$\theta=\PP(Y=1)$.
We take $20$ linearly-spaced 
$\theta \in (0.5, 1)$ and plot both $\Lambda_0$ (thick red segments) and
the reduced range $[\LBA^\star, \UBA^\star]$ (superimposed, thin blue lines). %
Decreasing 
$\theta$ from $1$ (top line at $\lamB=0.5$) to $0.5$ (bottom line at $\lamB=0$) yields an increase in $\LBA^\star$, thereby restricting the range of allowed $\MA$ models and fully specifying it for $\theta=0.5$ when~$Z:=Y$.

Extending the analytical treatment of~\cref{sec:examples} to more general settings is nontrivial.
To characterise the entailed constraints 
in generic settings,
we therefore resort to numerical simulations (see~\cref{app:experiments} for all technical details): 
we generate random instances of consistent $\PP_{XZ}$ and $\PP_{YZ}$, compute the space of solutions~$\Lambda_\Ccal$, and compare it to~$\Lambda_0$. 
A specific instance is shown in~\cref{fig:experiments}~\textbf{(b)};
see~\href{https://github.com/lgresele/structural-causal-marginal/blob/main/gifs/parameter_sweep_generic.gif}{[\underline{GIF1}]}~\href{https://github.com/lgresele/structural-causal-marginal/blob/main/gifs/parameter_sweep.gif}{[\underline{GIF2}]} for additional visualisations, where we fix a conditional $\PP_{Z|XY}$ and plot $\Lambda_\Ccal$ and $\Lambda_0$ for different choices of $\PP_X, \PP_Y$.\footnote{\label{footnote:cond}We fix a $\PP_{Z|XY}$ to ensure a solution to the statistical marginal problem exists; only the $\PP_{XZ}$ and $\PP_{YZ}$ derived from it are subsequently used to computate the solution spaces.}
The parameters used to generate~\cref{fig:experiments}~\textbf{(b)} violate some of the restrictive assumptions of~\cref{sec:examples} (most notably $X \indep Z$ and $P(Y=0,Z=1)=0$), 
and show that the schematic visualisation in~\cref{fig:overview} captures some aspects of the general case: the structural causal marginal problem can yield constraints for both marginal SCMs $\MA$ and $\MB$, and $\Lambda_0$ and $\Lambda_{\Ccal}$ are different. Moreover, we see that $(\lamA^{\text{max}}, \lamB^{\text{max}}) \in \Lambda_\Ccal$, consistent with~\cref{prop:lambda_max}.
In~\cref{fig:experiments}~\textbf{(c)}, we plot 
the cumulative distribution functions (CDFs) of the ratios between the blue and red areas (i.e., $(\UBA-\LBA)(\UBB-\LBB)/|\Lambda_0|$) in blue, and the ratio between the green and red areas (i.e., $|\Lambda_\Ccal|/|\Lambda_0|$) in green. 
The CDFs are estimated based on 1,000 independent samples of $\PP(Z=1|X=i, Y=j)\sim\text{Beta}(\alpha,\beta)$ for $i, j \in \{ 0, 1\}$ and $\PP(X=1),\PP(Y=1)\sim \text{U}[0,1]$.\textsuperscript{\ref{footnote:cond}} We compare two scenarios: $\alpha=\beta=1$, i.e., a Uniform prior, shown as solid lines; and $\alpha=\beta=0.5$, 
leading to more deterministic conditionals, shown as
dashed lines.
Across both scenarios, a reduction (i.e., ratios smaller than one) can be observed at least 30\% of the time. 
Whereas many times there is no or only a small reduction, we also sometimes (with positive probability) observe
quite substantial reductions of 50+\%.
Moreover, we find that $\alpha=\beta=0.5$ leads to larger reductions, suggesting that more deterministic (joint) conditionals may impose stronger constraints.
Finally, we remark that 
$(\lamA^{\text{max}}, \lamB^{\text{max}}) \in \Lambda_\Ccal$ indeed holds across all runs. 

\section{Limitations and Extensions}
\label{sec:extensions}
So far, we have focused on one of the simplest instances of the causal marginal problem involving (i) \textit{only two marginals}, each consisting of a (ii) \textit{Boolean} cause-effect model, under (iii) \textit{causal sufficiency}~(\cref{ass:causal_sufficiency}).
This simplified setting allowed us to %
focus on the main points %
and 
to visualise the problem in 2D~(see~\cref{fig:overview,fig:experiments}~\textbf{(b)}).
We now discuss how each of these restrictions can be relaxed to allow for more general settings, see~\cref{fig:extensions} for an illustration.

\paragraph{More than two marginals.}
Suppose that we have access to $m>2$ marginals, e.g., by separately observing  the effect of $(m-1)$ additional causes $Y_1, \dots, Y_{m-1}$ on $Z$ for the setting from~\cref{fig:teaser} and~\cref{sec:counterfactual_marginal_problem}.
Enforcing counterfactual consistency for each marginal would yield $m$ linear constraints in~\eqref{eq:constraints_matrix}, subject to which we could, e.g., solve for $[\LBA,\UBA]$.
While intuitively this should produce a tighter bound (if feasible), the number of response functions $h_k(X, Y_1, \dots, Y_{m-1})$ in the joint model grows as~$2^{2^m}$ (cf.~\eqref{eq:number_of_response_functions}) 
which  for large $m$ may pose computational challenges. 
In this case,
analogies to statistical learning theory~\citep{Vapnik} suggest restrictions on the capacity of
response functions %
which may render SCMs with~$m$ variables falsifiable from marginal observations with $k\ll m$ variables~\citep{overlapping}, see~\cref{app:learning_theory} for a  more detailed discussion.
\looseness-1 Causal graphs different from~\cref{fig:teaser} are, of course, also possible, leading to different parametrisations of the marginal and joint models; the general procedure of deriving linear constraints by enforcing counterfactual consistency and finding the corresponding feasible set would still apply. Note that causal modularity implies that (under~\cref{ass:causal_sufficiency}) the joint model can be specified by \textit{separately} describing the relations between each variable and its causal parents.
Hence, if each variable is observed jointly with \textit{all} its  parents in at least one marginal, this provides the same information as joint observation of all variables. This would for example be the case if we had the graph $X \leftarrow Z \leftarrow Y$ or $X \leftarrow Z \rightarrow Y$, but does not hold in the considered case of~\cref{fig:teaser}.

\paragraph{Dependent causes.}
In our example of \cref{fig:teaser}, if additionally we have dependent causes, e.g., $X\rightarrow Y$, our approach still applies but needs to be modified:
first, the \textit{joint} model 
now also involves a distribution over the four response functions  generating $Y$ from $X$;
second, the \textit{marginal} $Y\rightarrow Z$ model is now confounded by $X$, which requires specifying a joint distribution over $(Q_Y,Q_Z)$---see also~\cref{app:confounded_case}. %

\paragraph{Beyond Boolean variables.}
It is straightforward to extend our approach to arbitrary (non-binary) \textit{categorical} variables. 
As described in generality in~\cref{sec:background}, there would be more response functions, and the marginals may no longer be described by a single parameter but would still be constrained via~\eqref{eq:causal_Markov_kernel}. The projection operators~\eqref{eq:projection_operators} remain the same, and the constraints would be derived analogously to~\eqref{eq:constraints_scalar} with sums over the respective domains.
For \textit{continuous} variables, it is less clear how to proceed, as no simple parametrisation such as~\eqref{eq:response_function_framework} exists in general.
However, recent work suggests that assumptions on the allowed class of functions $f_i$ in~\eqref{eq:structural_equations} such as Lipschitz-continuity have non-trivial 
implications for partial identification~\cite{gunsilius2018non,gunsilius2019path}, see also~\cite{kilbertus2020class,zhang2021bounding} for recent progress. 
\looseness-1 An alternative is to discretise continuous variables, e.g., by  thresholding. %

\renewcommand{\yshift}{1.5em}
\renewcommand{\xshift}{3.5em}
\renewcommand{\xzcol}{cb-burgundy!60!white}
\renewcommand{\yzcol}{cb-blue!60!white}
\newcommand{\xycol}{cb-green-sea!80!white}
\begin{figure}
    \centering
    \begin{tikzpicture}
			\centering
			\node (X1) [obs,thick] {$V_1$};
			\node (X2) [obs, xshift=2*\xshift] {$V_2$};
			\node (X3) [obs, below=of X1, xshift=-\xshift,yshift=\yshift] {$V_3$};
			\node (X4) [obs, below=of X1, xshift=3*\xshift,yshift=\yshift] {$V_4$};
			\node (X5) [obs, below=of X3, xshift=2*\xshift,yshift=\yshift] {$V_5$};
 			\edge [thick] {X1}{X2};
 			\edge [thick] {X1}{X3};
			\edge [thick] {X1}{X5};
            \edge [thick] {X2}{X4};
			\edge [thick] {X3}{X5};
			\edge [thick] {X4}{X5};
			\path[<->,dashed, thick] (X1) edge[bend right=45] (X3);
			\path[<->,dashed, thick] (X2) edge[bend right=-45] (X4);
			\path[<->,dashed, thick] (X4) edge[bend right=-45] (X5);
 			\tikzset{plate caption/.style={caption, node distance=0, inner sep=3pt, above left=2pt and 5pt of #1.north,text height=0em,text depth=0em}}
 			\plate[inner sep=0.5em, ultra thick, color=\yzcol, yshift=3pt] {M1}{(X1) (X2)}{\textcolor{\yzcol}{\textbf{Dataset 1}}};
 			\tikzset{plate caption/.style={caption, node distance=0, inner sep=0pt, below left=-3pt and 45pt of #1.south,text height=0em,text depth=0em}}
 			\plate[inner sep=0.5em, ultra thick, color=\xzcol, yshift=3pt, xshift=-2pt] {M2}{(X3) (X4) (X5)}{\textcolor{\xzcol}{\textbf{Dataset 2}}};
 			\tikzset{plate caption/.style={caption, node distance=0, inner sep=0pt, above right=-10pt and 20pt of #1.north,text height=0em,text depth=0em}}
 			\plate[inner sep=0.5em, ultra thick, color=\xycol, yshift=-3pt, xshift=2pt] {M3}{(X2) (X4) (X5)}{\textcolor{\xycol}{\textbf{Dataset 3}}};
		\end{tikzpicture}
    \caption{%
    \textbf{Illustration of a More General Version of the Causal Marginal Problem.} Here, we have a causal graph over $n=5$ causal variables and observe $m=3$ marginals over the subsets $\Wb_1=\{V_1, V_2\}$, $\Wb_2=\{V_3, V_4, V_5\}$, and $\Wb_3=\{V_2, V_4, V_5\}$. Dashed bi-directed arrows indicate unobserved confounding.%
    }
    \label{fig:extensions}
\end{figure}
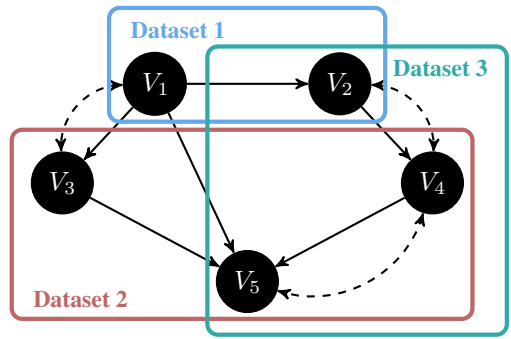

\paragraph{Unobserved confounding.}
When~\cref{ass:causal_sufficiency} is violated, the Markov factorisation~\eqref{eq:Markov_factorisation} does not hold and we cannot consider the distributions $\PP_{R_i}$ of each response function variable separately. Instead, we need to parametrise their joint distribution~$\PP_\Rb$ (which can drastically increase the number of parameters) and derive the constraints imposed by the observational distributions via~\eqref{eq:observational_distribution} instead of~\eqref{eq:causal_Markov_kernel}.\footnote{For known confounding structures, a partially factorised $\PP_\Rb$ could be used, but this typically leads to nonlinear constraints.}
With hidden confounders, a gap remains between the first and the second rung even when the causal graph is known: interventional distributions are no longer determined by the observed $\PP_\Vb$ as~\eqref{eq:truncated_factorisation} no longer holds.
Knowing some of the do-probabilities from experimental data therefore provides additional information and imposes further constraints via~\eqref{eq:interventional_distribution}.
For a more detailed treatment of a confounded version of the setting from~\cref{sec:counterfactual_marginal_problem}, we refer to~\cref{app:confounded_case}.
Finally, we note that in confounded settings it could also be interesting to consider an instantiation of the causal marginal problem based on interventional models such as causal Bayesian networks~\citep[CBNs;][]{Spirtes1993} instead of SCMs.

With these extensions in mind, we finally give a more general definition of the causal marginal problem~(cf.~\cref{fig:extensions}).
\vspace{-0.75em}
\begin{definition}[Causal marginal problem]
	Consider $m$ marginal causal (interventional or counterfactual) models $\Mcal_1, \ldots, \Mcal_m$ over distinct but overlapping sets of variables $\Wb_1, \ldots, \Wb_m \subseteq \{ V_1, \ldots, V_n \}$, respectively.
    The \textit{causal marginal problem} consists of determining the space of %
	joint causal (interventional or counterfactual) models  $\Mcal$ over $\Wb_1 \cup \ldots \cup \Wb_m$ which are (interventionally or counterfactually) consistent with the marginal ones.
	\label{def:causal_marg_prob}
\end{definition}
\section{Related Work}
\label{sec:discussion}
\looseness-1 The problem of merging causal models
involving overlapping subsets of variables has also been considered by~\citet{overlapping, mejia2021obtaining},\footnote{Federated learning \citep{kairouz2021advances} is loosely related: there, the aim is to 
learn from
data from multiple sources (clients), 
and each client's data is accessible and processed \textit{locally}, whereas in our setting  all marginals are available and processed {\em globally}.} though focusing on interventional (rung 2) quantities.
Marginalisation of SCMs, i.e., the inverse problem of merging, has been discussed by~\citet{bongers2016structural} and~\citet{Rubensteinetal17}. %
The latter introduce a notion of \textit{interventional} consistency between marginal and joint SCMs, which complements ours of \textit{counterfactual} consistency%
~(\cref{def:counterfactual_consistency}).
\looseness-1 A related type of consistency between different abstractions of the same underlying causal system has been studied by~\citet{chalupka2016multi,chalupka2017causal,beckers2019abstracting,beckers2020approximate}.

In the present work, we have explored the implications of merging for the space of allowed marginal and joint models, i.e., \textit{partial identification of SCMs}.
A parallel literature instead aims to identify \textit{specific causal queries} from a given collection of observational and experimental datasets (involving subsets of variables), a task referred to as \textit{transportability} or \textit{data fusion}~\cite{pearl2014external,bareinboim2016causal}---see also~\citet{chau2021bayesimp} for uncertainty quantification in this context and~\citet{leecausal} for a combination with proxy-based approaches.

\looseness-1 Both our and the aforementioned line of work assume that the causal graph is known a priori.
For \textit{causal structure learning} approaches for the setting of multiple datasets involving overlapping sets of variables, we refer to~\citet{triantafillou2010learning, tillman2014learning,triantafillou2015constraint, huang2020causal}.%
\section{Discussion}
\paragraph{Empirical content of counterfactuals.}
The use of counterfactuals in causal inference %
has long been a subject of debate;
as %
summarised by%
~\citet{shafer1996art}:
  ``\textit{were counterfactuals to have objective meaning, we might take them as basic, and define probability and causality in terms of them}''. %
Some prominent approaches to causal inference indeed regard counterfactuals (or potential outcomes) as foundational~\cite{Imbens2015, pearl2009causality}.
Others question the legitimacy of %
models
allowing for direct formulation of counterfactual 
queries (such as SCMs):
\citet{dawid2000causal} 
terms them {\em ``metaphysical''}, arguing that they either yield
unscientific (i.e., empirically irrefutable) statements or are unnecessary in that the inferences for which they are used could also be rephrased in non-counterfactual terms. 
Our work
illustrates a possible mode of 
falsifiability for counterfactual models: {some SCMs may be falsified when 
previously unobserved variables become observable together with subsets of the original ones (e.g., through a new experiment or study) and are consistently merged into a joint model.
For example, %
in the setting of~\cref{fig:teaser}, the exogenous variable associated to $Z$ could (partly) correspond to~$Y$: together with
$Y\to Z$, observing $\PP_{YZ}$  would then provide (partial) information on what is otherwise unobserved.
This reflects a view according to which counterfactuals do carry an empirical message and {\em ``may earn predictive power''}
when {\em ``the uncertainty-producing variables offer the potential of being observed sometime in the future (before our next prediction or action)''}~\citep[][\S~7.2.2]{pearl2009causality}. We further illustrate this point with an example in~\cref{app:illustrative_example}.
\vspace{-0.1em}
Another insight %
is that {\em interventional marginal  models} (i.e., a causal graph and corresponding observational distribution) {\em can entail constraints for counterfactual ones} such as SCMs. 
In other words, questions regarding model consistency may also be meaningful when marginal and joint models do not refer to the same rung in the ladder of causation. %
This intertwines the two model classes (rungs two and three).
%
%
%
%
%
%
    %
    %
    %
    %
%
    %
    %
        %
%
%
    %
    %
    %
    %
%
%
%
%
%
%
%
%
%
%
%
%
%
%
%
%
%
%
%
%
%
%
%
%

%
%
%
%
%
%
%
%

%
%
%
%
\paragraph{SCMs and falsifiability.} \citet{popper1989logik} considers falsifiability a crucial property of {\it scientific} hypotheses: %
unfalsifiable ones belong to the realm of metaphysics,
and falsifiable ones
are increasingly corroborated
as many attempts to falsify them fail. %
\cref{prop:lambda_max} suggests that some SCMs are intrinsically `harder' %
to falsify %
in that the space of interventional models %
they can be consistently merged with is larger%
: %
those marginal models with the weakest counterfactual influence can always be consistently merged and 
are thus not falsifiable through additional variables. 
Conversely, if a marginal %
SCM with a strong counterfactual influence can (repeatedly)
be merged consistently with new marginals, %
we obtain indirect evidence for it in the Popperian sense. %
\textit{Which classes of causal models (beyond our Boolean setting) offer a larger space of possible falsifications?}
This parallels the idea of capacity measures in supervised learning, where the generalisation gap is provably smaller if a class of allowed explanations has small capacity relative to the dataset size. The latter means that the space of datasets that would falsify it (in that they cannot be fitted by any explanation in the class) is large~\citep{CorSchVap}. By analogy, this would suggest to prefer SCMs that are easier to falsify: the question may be investigated in future work as a first step towards a 
`statistical learning theory of causal data fusion'%
.
}
%
%
%
%
%
%
%
    %
    
%
%
%
%
%
%
%

%
%
%
%

%
%

%
%
%
%
%
%

%

%
\paragraph{Concluding remarks.}
\label{sec:conclusion}
We introduced the structural causal marginal problem as a framework for
merging causal information from different datasets. 
While previous work focused on bounds on counterfactuals from \textit{joint} observations, we have emphasised bounds and falsifiabiliy that come from {\it marginal} causal information involving different subsets of variables. \looseness-1 
This way, causal insights emerge from `bringing puzzle pieces together' rather than from complete datasets.

\section*{Software and Data}
Code is available at \href{https://github.com/lgresele/structural-causal-marginal}{https://github.com/lgresele/structural-causal-marginal}.

\section*{Acknowledgements}
\looseness-1
We thank Sergio Hernan Garrido Mejia, Claudia Shi, Shiva Kasiviswanathan, Filippo Camilloni, Krikamol Muandet, Sander Beckers, Armin Keki\'c, Atalanti Mastakouri and Kailash Budhathoki for valuable discussions; and the anonymous reviewers for helpful comments.
This work was supported by the German Federal Ministry of Education and Research (BMBF): Tübingen AI Center, FKZ: 01IS18039A, 01IS18039B; and by the Machine Learning Cluster of Excellence, EXC number 2064/1 – Project number 390727645.

\bibliography{references.bib}
\bibliographystyle{icml2021}

\appendix
\onecolumn

\section{Parametrisation of Boolean Causally-Sufficient Cause-Effect Models}
\label{app:single_param_scm_s}
The definition of the response functions of the joint model is provided in~\cref{tab:15functions}.
We now derive the family of SCMs that are (interventionally) consistent with $X\to Z$ and the observed $\PP_{XZ}$ for the setting considered in~\cref{sec:Boolean_cause_effect}, i.e., assuming causal sufficiency and Boolean variables. 

First, we note that the marginal distribution $\PP_X$ completely determines the distribution of $R_X$ in~\eqref{eq:MA}.

Next, we consider how $\PP_{Z|X}$ constrains the distribution of $R_Z$, i.e., the probability vector $\ab\in\Delta^3$.

From~\eqref{eq:causal_Markov_kernel} and the definition of the response functions $f_i$, we obtain the following two independent constraints:
\begin{align}
\label{eq:marginal_constraint1}
    \PP_{Z|X}(Z=0|X=0)
    =:
    p_{00}%
    =
    a_0+a_2\\
\label{eq:marginal_constraint2}
    \PP_{Z|X}(Z=0|X=1)
    =:
    p_{01}%
    =
    a_0+a_3
\end{align}

Additionally, we have the simplex constraint:
\begin{equation}
\label{eq:marginal_simplex_constraint}
1=a_0+a_1+a_2+a_3  
\end{equation}

We now solve the under-determined system of equations \eqref{eq:marginal_constraint1}, \eqref{eq:marginal_constraint2}, \eqref{eq:marginal_simplex_constraint} by setting $a_0=:\lamA$.

From~\eqref{eq:marginal_constraint1} and~\eqref{eq:marginal_constraint2}, this yields
\begin{align}
a_2&=p_{00}-\lamA\\
a_3&=p_{01}-\lamA
\end{align}
and finally, by substitution in~\eqref{eq:marginal_simplex_constraint},
\begin{equation}
    a_1=1-\lamA-(p_{00}-\lamA)-(p_{01}-\lamA)=1-p_{00}-p_{01}+\lamA
\end{equation}

Writing this as a vector, we obtain the following form for~$\ab$:
\begin{equation}
\label{eq:SCM_A_characterisation}
    \ab
    =
    \begin{pmatrix}
    0\\
    1-p_{00}-p_{01}\\
    p_{00}\\
    p_{01}
    \end{pmatrix}
    +
    \lamA
    \begin{pmatrix}
    1\\
    1\\
    -1\\
    -1
    \end{pmatrix}
\end{equation}
with a single free parameter $\lamA\in\RR$. 

Since we require $0\leq a_i\leq 1, \forall i$ for $\ab$ to be a valid probability vector, we find the admissible range of $\lamA$ to be:
\begin{equation}
\label{eq:lambda_A}
    \max\left\{0, p_{00}+p_{01}-1\right\}
    \leq 
    \lamA
    \leq
    \min\left\{p_{00},p_{01}\right\}
\end{equation}

Similarly, we can characterise the other marginal SCM $\Mcal^B$ over $Y\rightarrow Z$ in terms of its observational distribution.
Denoting $p'_{ij}:=\PP(Z=i|Y=j)$, this yields analogously:
\begin{equation}
\label{eq:SCM_B_characterisation}
    \bb
    =
    \begin{pmatrix}
    0\\
    1-p'_{00}-p'_{01}\\
    p'_{00}\\
    p'_{01}
    \end{pmatrix}
    +
    \lamB
    \begin{pmatrix}
    1\\
    1\\
    -1\\
    -1
    \end{pmatrix}
\end{equation}
with
\begin{equation}
\label{eq:lambda_B}
    \max\left\{0, p'_{00}+p'_{01}-1\right\}
    \leq 
    \lamB
    \leq
    \min\left\{p'_{00},p'_{01}\right\}
\end{equation}

\newpage
\section{Parametrisation of the Joint SCM with two Boolean causes}
\label{app:joint_model_parametrisation}
\begin{table}[h!]
    \centering
    \caption{Definition of the 16 response functions $h_{k}(X,Y)$ from~\eqref{eq:MC} mapping two Boolean inputs to a Boolean output. Each row corresponds to one of the four different combinations $(x,y)$ of the two inputs $X$ and $Y$; columns correspond to different $h_k$; and cells indicate the corresponding output of $h_k(x,y)$.
    }
    \label{tab:15functions}
    \begin{tabular}{cc|cccccccccccccccc}
		\toprule $X$& $Y$ & $h_{0}$ & $h_{1}$ & $h_{2}$ & $h_{3}$ & $h_{4}$ & $h_{5}$ & $h_{6}$ & $h_{7}$ & $h_{8}$ & $h_{9}$ & $h_{10}$ & $h_{11}$ & $h_{12}$ & $h_{13}$ & $h_{14}$ & $h_{15}$ \\
		\midrule $0 $& $ 0$ & 0 & \rone & 0 & \rone & 0 & \rone& 0 & \rone & 0 & \rone & 0 & \rone & 0 & \rone & 0 & \rone \\
		$0 $& $ 1 $ & 0 & 0 & \rone & \rone & 0 & 0 & \rone & \rone & 0 & 0 & \rone & \rone & 0 & 0 & \rone & \rone \\
		1& $ 0$ & 0 & 0 & 0 & 0 & \rone & \rone & \rone & \rone & 0 & 0 & 0 & 0 & \rone & \rone & \rone & \rone \\
		1& 1 & 0 & 0 & 0 & 0 & 0 & 0 & 0 & 0 & \rone & \rone & \rone & \rone & \rone & \rone & \rone & \rone \\
		\bottomrule
	\end{tabular}
\end{table}

\section{Details for the Example from~\cref{sec:examples}}
\label{app:examples}
We now provide a more detailed account of the worked-out example from~\cref{sec:examples}.

Recall that we make the following assumptions:
\begin{enumerate}[label=(\roman*)]
    \item
    $X\indep Z$%
    \item
    $\PP(Y=0,Z=1)=0$,
    \item
    $\PP(Z=1)
    =0.5$,
    \item
    $0<\PP(X=0),\PP(Y=0)<1$.
\end{enumerate}

\subsection{Derivation of the SCM $\MB$ over $Y\to Z$}
Recall from~\cref{sec:Boolean_cause_effect} that the SCM $\MB$ over $Y\to Z$ is characterised by the probability vector
\begin{equation*}
    \bb
    =
    \begin{pmatrix}
    0\\
    1-p'_{00}-p'_{01}\\
    p'_{00}\\
    p'_{01}
    \end{pmatrix}
    +
    \lamB
    \begin{pmatrix}
    1\\
    1\\
    -1\\
    -1
    \end{pmatrix}
\end{equation*}
where $p'_{ij}=\PP(Z=i|Y=j)$, see~\cref{app:single_param_scm_s}.

Now by assumption (ii) and~\eqref{eq:causal_Markov_kernel}, we have that
\begin{equation}
    \PP(Z=1|Y=0)=0=b_1+b_3
\end{equation}
from which we conclude that
\begin{equation}
    b_3=p'_{01}-\lamB=0 \Leftrightarrow \lamB=p'_{01}
\end{equation}
and
\begin{equation}
    b_1=1-p'_{00}-p'_{01}+\lamB=0 \Leftrightarrow p'_{00}=1.
\end{equation}

This yields the following intermediate form of $\bb$:
\begin{equation*}
    \bb
    =
    \begin{pmatrix}
    p'_{01}\\
    0\\
    1-p'_{01}\\
   0
    \end{pmatrix}
\end{equation*}

Next, by assumptions (ii) and (iii) we have that
\begin{equation}
    \PP(Z=1)=0.5=\PP(Y=1)\PP(Z=1|Y=1)=\PP(Y=1)(1- p'_{01})
\end{equation}
Writing $\theta:=\PP(Y=1)$ and solving for $p'_{01}$ we find
\begin{equation}
    p'_{01}=\frac{2\theta -1}{2\theta}
\end{equation}
which yields the final expression $\bb=(\frac{2\theta -1}{2\theta},0,\frac{1}{2\theta},0)^\top$ as stated in the main paper.

In other words, $Z:=Y$ with probability $\frac{1}{2\theta}$ and $Z:=0$ otherwise, which corresponds to the AND model from~\eqref{eq:example_and_model}.

\subsection{Derivation of the family of SCMs $\MA$ over $X\to Z$}
Next, we consider the family of SCMs $\MA$ over $X\to Z$ which is characterised by the response function probability vector
\begin{equation*}
    \ab
    =
    \begin{pmatrix}
    0\\
    1-p_{00}-p_{01}\\
    p_{00}\\
    p_{01}
    \end{pmatrix}
    +
    \lamA
    \begin{pmatrix}
    1\\
    1\\
    -1\\
    -1
    \end{pmatrix}
\end{equation*}
where $p_{ij}=\PP(Z=i|X=j)$, see~\cref{app:single_param_scm_s}.

By assumption (iii) and independence of $X$ and $Z$ (assumption (i)), we have that:
\begin{equation}
    \PP(Z=0)=0.5=p_{00}=p_{01}
\end{equation}

Substituting the above into the expression for $\ab(\lamA)$, we obtain:
\begin{equation}
\label{eq:SCM_A_characterisation}
    \ab(\lamA)
    =
    \begin{pmatrix}
    \lamA\\
    \lamA\\
    0.5-\lamA\\
    0.5-\lamA
    \end{pmatrix}.
\end{equation}
as well as $\lamA\in[\lamA^\text{min},\lamA^\text{max}]=[0,0.5]$
as stated in~\cref{sec:examples}.

Moreover, for $\lamA=0.5$, we have that $Z:=0$ or $Z:=1$, both with probability $0.5$ corresponding to~\eqref{eq:example_indepedent_model}, whereas for $\lamA=0$, we have that $Z:=X$ or $Z:=1-X$, both with probability $0.5$ corresponding to~\eqref{eq:example_xor_model}.

\subsection{Enforcing counterfactual consistency between the marginal SCMs}
We now explore the implications of enforcing counterfactual consistency between the two marginal SCMs taking the forms derived in the previous two subsections.
To this end, we consider what the valid choices for the joint model, i.e., for $\cb$, are, and whether this imposes additional constraints on the marginal models. 

First, note that for $\MB$ we have $b_1=b_3=0$. According to~\eqref{eq:constraints_scalar}, this implies:
\begin{align}
 b_1&=0=\PP(X=0) ( c_{3} + c_{7} + c_{11} + c_{15}) + \PP(X=1) (c_{12} + c_{13} + c_{14} + c_{15}) \\
 b_3&=0=	\PP(X=0) (c_{1} + c_{5} + c_{9} + c_{13}) + \PP(X=1) ( c_{4} + c_{5} + c_{6} + c_{7})  
\end{align}
Together with $0<\PP(X=0)<1$ from assumption (iv), and since $c_i\geq 0$, we must have that:
\begin{equation}
	c_{1} = c_{3} = c_{4} = c_{5} = c_{6} = c_{7} = c_{9} = c_{11} = c_{12} = c_{13} = c_{14} = c_{15} = 0. \label{eq:impossible_functions}
\end{equation}
This only leaves $c_0, c_2, c_8, c_{10}$ as non-zero elements of $\cb$. 

We now consider counterfactual consistency with $\MA$.
Writing the constraint $\ab(\lamA)=\Ab\cb$ from~\eqref{eq:constraints_matrix} subject to~\eqref{eq:impossible_functions}, we obtain:
\begin{equation}
    \begin{pmatrix}
    \lamA\\
    \lamA\\
    0.5-\lamA\\
    0.5-\lamA
    \end{pmatrix}
    =
    \begin{pmatrix} 
		1 & \PP(Y=0)  & \PP(Y=0) & \PP(Y=0)     \\ 
		0 & 0 & 0 & \PP(Y=1)   \\ 
		0 & 0 & \PP(Y=1)  & 0    \\ 
		0 & \PP(Y=1)  & 0  & 0    \\       
	\end{pmatrix} 
	\begin{pmatrix}            
		c_{0} \\         
		c_{2}     \\  
		c_{8} \\          
		c_{10} \\   
	\end{pmatrix}
\end{equation}

Whereas $(c_2, c_8, c_{10})=\frac{1}{\PP(Y=1)}(\lamA, 0.5-\lamA, 0.5-\lamA)$ are always valid probabilities for $\lamA\in[0,0.5]$ since we must have $\theta=\PP(Y=1)\geq 0.5$ by assumptions (ii) and (iii) (see footnote 7 in~\cref{sec:examples}), solving for $c_0$ yields:
\begin{align}
    &c_0=\lamA -\PP(Y=0)(c_2+c_8+c_{10})=\lamA - \frac{\PP(Y=0)}{\PP(Y=1)}(1-\lamA)\geq0\\
    \Leftrightarrow 
    &\lamA\left(1+\frac{1-\theta}{\theta}\right)\geq \frac{1-\theta}{\theta}
 \Leftrightarrow \lamA\geq 1-\theta
\end{align}
That is, in order for there to be a valid solution $\cb$, the additional constraint $\lamA\geq 1-\theta$ must be satisfied.

\section{Structural Causal Marginal Problem \textit{With Unobserved Confounding}}
\label{app:confounded_case}
We now provide a more detailed treatment of a version of the structural causal marginal problem from~\cref{sec:counterfactual_marginal_problem} in which causal sufficiency~(\cref{ass:causal_sufficiency}) is violated, i.e., allowing for arbitrary unobserved confounding. 
Specifically, we consider the setting in which both marginal SCMs $X\rightarrow Z$ and $Y\rightarrow Z$ are confounded (i.e., there exist unobserved variables influencing $\{X,Z\}$ and $\{Y,Z\}$, respectively).
Likewise, the joint model $\{X,Y\}\rightarrow Z$ is also assumed to be potentially confounded.

On a technical level, unobserved confounding 
manifests in a potential \textit{dependence of the exogenous noise terms} in the structural equations. 
In other words, the distribution over exogenous variables no longer factorises---in contrast to the unconfounded case.
Within the response function framework, this means that the input or cause is no longer independent of the function or mechanism generating the effect from the cause(s).
This means that we cannot parametrise the cause distribution and distribution over functions separately, but instead need to consider their joint distribution~$\PP_\Rb$.

\subsection{Constraints imposed on the marginal SCMs by the observational marginal distributions}
First we consider the marginal SCMs $\MA$ and $\MB$ (defined as in the main paper) and investigate how they are constrained by the observed $\PP_{XZ}$ and $\PP_{YZ}$.

For notational convenience, we denote the observational distributions by
\begin{equation}
\begin{aligned}
    \PP(X=i,Z=j)&=\alpha_{ij},
    \\
    \PP(Y=i,Z=j)&=\beta_{ij},
\end{aligned}
\end{equation}
for $i,j\in\{0,1\}$,
and collect them in vectors $\alphab,\betab\in\Delta^3\subseteq[0,1]^4$.

Similarly, we parametrise the joint distributions over the corresponding response function variables as follows:
\begin{equation}
\begin{aligned}
    \PP(R_X=i, R_Z=j)&=\PP(X=i, R_Z=j)=\qA_{ij}, 
   \\
   \PP(Q_Y=i, Q_Z=j)&=\PP(Y=i, Q_Z=j)=\qB_{ij}, 
\end{aligned}
\end{equation}
for $i\in\{0,1\}, j\in\{0,1,2,3\}$,
and collect them in vectors $\qbA,\qbB\in\Delta^7\subseteq[0,1]^8$.

Since the Markov factorisation~\eqref{eq:Markov_factorisation} does not hold under hidden confounding, we need to derive the constraints imposed by the observational distributions $\PP_{XZ}$ and $\PP_{YZ}$ via~\eqref{eq:observational_distribution} instead of~\eqref{eq:causal_Markov_kernel}. This yields:
\begin{equation}
\begin{aligned}
\alpha_{ij}=\PP(X=i,Z=j)
    &=
    \sum_{i'=0}^1\sum_{j'=0}^3 \PP(R_X=i',R_Z=j')\II\{i=i'\}\II\{j=f_{j'}(i')\}=\sum_{j'=0}^3 \qA_{ij'} \II\{j=f_{j'}(i)\}
    \\
        \beta_{ij}=\PP(Y=i,Z=j)
    &=
    \sum_{i'=0}^1\sum_{j'=0}^3 \PP(Q_Y=i',Q_Z=j')\II\{i=i'\}\II\{j=f_{j'}(i')\}=\sum_{j'=0}^3 \qB_{ij'} \II\{j=f_{j'}(i)\}
\end{aligned}
\end{equation}
for $i,j\in\{0,1\}$.
Writing the above in matrix form, we thus obtain the constraints
\begin{equation}
\begin{aligned}
\alphab&=\LbA\qbA
    \\
    \betab&=\LbB\qbB
\end{aligned}
\end{equation}
where $\LbA,\LbB\in\{0,1\}^{4\times 8}$ are binary constraint matrices.

We conclude that the space of all potentially confounded SCMs $\MA$ over binary $X\rightarrow Z$ consistent with a given observational joint distribution $\alphab$ is parametrised by all $\qbA\in\Delta^7$ which satisfy $\alphab=\LbA\qbA$.

Unlike in the unconfounded case, this results in four free parameters: seven free parameters with three linearly independent constraints. (Note that matching the observational distribution only eliminates three instead of four free parameters since the fourth constraint is a linear combination of the other three: $\alpha_{11}=1-\alpha_{00}- \alpha_{01} -\alpha_{10}$.)

Analogously, any potentially confounded SCM $\MB$ over binary $Y\rightarrow Z$ consistent with a given observational joint distribution $\betab$ is parametrised by all $\qbB\in\Delta^7$ which satisfy $\betab=\LbB\qbB$. 

\subsection{Additional constraints imposed via experimental data}
Whereas in the unconfounded cases, the observational $\PP_{Z|X}$ and interventional $\PP_{Z|do(X)}$ conditionals are identical,
$$\forall (x,z): \qquad \PP(Z=z|X=x)=\PP(Z=z|do(X=x)),$$
this is not the case when unobserved confounding is allowed: there may exist some $(x,z)$ such that
$$\PP(Z=z|X=x)\neq \PP(Z=z|do(X=x)).$$
Intuitively, with hidden confounding, the observational conditional captures two types of dependence: (i) the direct dependence between $X$ and $Z$, and (ii) the (indirect) dependence due to their (unobserved) common cause. 
The interventional distribution, on the other hand, only comprises the first type (i).
As a consequence, having access not only to the marginal observational distribution $\PP_{XZ}$ but also to the do probabilities $\PP(Z=z|do(X=x))$ may impose additional constraints. 

Specifically, we have the following additional constraints:
\begin{equation}
\begin{aligned}
\alpha^\textsc{IV}_{ij}&=\PP(Z=j|do(X=i))=\sum_{j'=0}^3 (\qA_{0j'}+\qA_{1j'}) \II\{j=f_{j'}(i)\}
\\
\beta^\textsc{IV}_{ij}&=\PP(Z=j|do(Y=i))=\sum_{j'=0}^3 (\qB_{0j'}+\qB_{1j'}) \II\{j=f_{j'}(i)\}
\end{aligned}
\end{equation}
Note that in contrast to before, we are additionally summing over the first subscript of $q$ leading to the $(q_{0j'}^A+q_{1j'}^A)$ terms. This is because---unlike in the observational case---the value of the exogenous variable $R_X$ associated with $X$ does not matter, since $X$ is fixed by intervention, rather than taking on its natural value through the mechanism $f$.

The above can be written in matrix form as follows:
\begin{equation}
    \begin{aligned}
    \alphab^\textsc{IV}&=\LbA_\textsc{IV}\qbA
    \\
\betab^\textsc{IV}&=\LbB_\textsc{IV}\qbB.
    \end{aligned}
\end{equation}

If experimental data in the form of $\alphab^\textsc{IV},\betab^\textsc{IV}$ (or parts thereof) are available, we can use it to additionally constrain $\qbA,\qbB$. The number of free parameters for each of $\qbA,\qbB$ can be reduced by at most two more this way, leaving a total of two free parameters each.

\subsection{Additional constraints via assumptions such as monotonicity}
Another way of reducing the number of free parameters is by means of additional assumptions such as the \textit{monotonicity} assumption, which is common in epidemiology and economics, particularly in the context of instrumental variable (IV) models~\cite{imbens1994identification},  and posits that  there are no ``defiers'', i.e., the weight of the NOT function $f_3$ is zero:
\begin{equation}
    \begin{aligned}
    \PP(R_Z=3)&=0=\qA_{03}+\qA_{13}
\\
\PP(Q_Z=3)&=0=\qB_{03}+\qB_{13}
    \end{aligned}
\end{equation}

\subsection{Parametrisation of the joint SCM}
Next, we parametrise the joint SCM~$\MC$ over $\{X,Y\}\rightarrow Z$.
Whereas in the unconfounded case, we were able to conclude that $X\independent Y$ (for otherwise one of the marginal SCMs would be confounded), this is not necessarily true in the more general confounded case.
Here, we will work under the assumption that we do not know the causal ordering between $X$ and $Y$ and therefore cannot specify a full SCM without additional assumptions or background knowledge.
We therefore proceed with specifying a \textit{partial} causal model, consisting of
 (i)   a joint distribution $\PP_{XY}$, and (ii)
    the structural equation generating $Z$
    \begin{equation}
        Z:=h_{S}(X,Y)
    \end{equation}

Note that such a model will only allow us to reason interventionally and counterfactually about joint interventions of the form $do(X:=x,Y:=y)$ but not about single node interventions, $do(X:=x)$ or $do(Y:=y)$, since we are not modelling the causal relationship between $X$ and $Y$. (And, since we never observe $X$ and $Y$ jointly, we may not be able to infer it.)

As in the unconfounded case, the response function variable $S$ takes values in $\{0, 1, ..., 15\}$ indexing the 16 functions $h_k:\{0,1\}^2\rightarrow \{0,1\}$ listed in~\cref{tab:15functions} in~\cref{app:joint_model_parametrisation}.

We parametrise this joint partial causal model over binary, potentially confounded $\{X,Y\}\rightarrow Z$ as follows:
\begin{equation}
    \PP(X=i,Y=j,S=k)=q_{ijk}
\end{equation}
and collect these probabilities of the $2\times 2 \times 16 =64$ joint states in a vector $\qb\in\Delta^{63}\subseteq[0,1]^{64}$.

\subsection{Enforcing consistency between the joint and marginal models}
We now impose the additional constraint that the two marginal SCMs $\MA$ and $\MB$ parametrised by $\qbA$ and~$\qbB$, respectively, must be counterfactually consistent (at the level of counterfactual involving $Z$ under changes to $X$ and $Y$) with the (partial) joint model parametrised by $\qb$.
To this end, we proceed as in the unconfounded case making use of the projection operators $\Pcal^Y_y$ and $\Pcal^X_x$ from~\eqref{eq:projection_operators} which given a particular value $Y=y$ or $X=x$ map the functions $h_k$ to functions $f_j(X)$ or $f_{j'}(Y)$, respectively. 

Specifically, for enforcing consistency of the joint model $\MC$ with $\MA$ over $X\rightarrow Z$ after marginalisation of $Y$, we obtain for all $i\in\{0,1\}$ and $j\in\{0,1,2,3\}$:
\begin{align}
    \qA_{ij}=\PP(X=i, R_Z=j)=\sum_{y=0}^1\,\,\sum_{k:\Pcal^Y_y(h_k)=f_j}\PP(X=i,Y=y,S=k)=\sum_{y=0}^1\sum_{k:\Pcal^Y_y(h_k)=f_j} q_{iyk}
\end{align}
Similarly, for consistency of $\MC$  with $\MB$  after marginalisation of $X$ we obtain for all $i\in\{0,1\}$ and $j\in\{0,1,2,3\}$:
\begin{align}
    \qB_{ij}=\PP(Y=i, Q_Z=j)=\sum_{x=0}^1\,\,\sum_{k:\Pcal^X_x(h_k)=f_j}\PP(X=x,Y=i,S=k)=\sum_{x=0}^1\sum_{k:\Pcal^X_x(h_k)=f_j} q_{xik}
\end{align}
This can be written in matrix form as
\begin{equation}
\begin{aligned}
    \qbA&=\KbA\qb
    \\
    \qbB&=\KbB\qb
\end{aligned}
\end{equation}
where $\KbA,\KbB\in\{0,1\}^{8\times 64}$ are binary constraint matrices.

\subsection{Linear program and polytope of solutions}
We can now reason about different types of interventional and counterfactual queries that can be expressed in terms of $\qbA$, $\qbB$, or $\qb$ subject to the constraints imposed by enforcing consistency:
\begin{enumerate}[label=(\roman*)]
\item  between the two families of marginal SCMs $\MA$ and $\MB$ parametrised by $\qbA$ and $\qbB$ with their respective observational distributions $\alphab$ and $\betab$; 
\item between the joint (partial) SCM $\MC$ parametrised by $\qb$ with the two marginal SCMs $\MA$ and $\MB$.
\end{enumerate}
Denoting the query of interest by $\Qcal$, this leads to the following optimisation problem which is again a linear program:
\begin{equation}\label{eq:LP_confounded}
    \begin{aligned}
    \minmax_{\qbA,\qbB\in\Delta^7, \qb\in\Delta^{63}} \quad \quad &\Qcal(\qbA,\qbB, \qb)\\
    \text{subject to:} \quad \quad &\alphab=\LbA\qbA\\
    &\betab=\LbB\qbB\\
    &\alphab^\textsc{IV}=\LbA_\textsc{IV}\qbA\\
    &\betab^\textsc{IV}=\LbB_\textsc{IV}\qbB\\
    &\qbA=\KbA\qb\\
    &\qbB=\KbB\qb
    \end{aligned}
\end{equation}
where $\alphab, \betab$ and $\alphab^\textsc{IV}, \betab^\textsc{IV}$ (provided experimental data is available)  are known constant probabilities, and $\LbA,\LbB, \LbA_\textsc{IV}, \LbB_\textsc{IV}, \KbA,\KbB$ are known constant, binary constraint matrices.

Similarly to~\eqref{eq:C_polytope} (see \cref{app:reformulation_linear_program} for more details) we can also first define the space of  allowed joint SCMs as a polytope:
\begin{align*}
    \mathcal{C}_\text{conf} := \left\{\qb \in \Delta^{63} \mid
    \exists \qbA,\qbB\in\Delta^7: \,\,
    \alphab=\LbA\qbA, \betab=\LbB\qbB, \alphab^\textsc{IV}=\LbA_\textsc{IV}\qbA, \betab^\textsc{IV}=\LbB_\textsc{IV}\qbB, \qbA=\KbA\qb, \qbB=\KbB\qb \right\}.
\end{align*}
The vertices of $\mathcal{C}_\text{conf}$ can be found using numerical solvers in analogy to how this is done in the unconfounded case, see \cref{app:reformulation_linear_program}.
We could then optimise queries over allowed $\qb$ simply by optimising over $\mathcal{C}_\text{conf}$, an equivalent formulation to \eqref{eq:LP_confounded}. Arguably, however, if one solely cares about a specific query, solving \eqref{eq:LP_confounded} is more direct.
Furthermore, if one is interested in the consistent marginal models, we can compute the projection of the vertices of $\mathcal{C}_\text{conf}$ onto, say, $\qbA$ via $\qbA=\KbA\qb$ and define the consistent SCMs from $X\to Z$ as the convex hull of the projected vertices.

\subsection{What counterfactual queries can be addressed by which model?}
We may wonder what types of counterfactual queries each of the marginal and joint SCMs may be able to answer, especially given that we only considered a partial specification of the joint SCM.
We summarise this as follows:
\begin{itemize}[leftmargin=5em]
    \item[$\MA$:] $\PP(Z_{do(x)}|x',z')$
    \item[$\MB$:] $\PP(Z_{do(y)}|y',z')$
    \item[$\MC$:] $\PP(Z_{do(x,y)}|x', y',z')$
    \item[None:] $\PP(Z_{do(x)}|x',y',z'), \PP(Z_{do(y)}|x',y',z')$
\end{itemize}

Answering the last type of query would require either additional assumptions such as $X\independent Y$, or knowledge of the qualitative causal relationship between $X$ and $Y$, either whether we have $X\to Y$ or $Y\to X$.

\section{Proof of \cref{prop:lambda_max}}
\label{app:lambda_max}
\newtheorem*{prop_restated}{Proposition \ref{prop:lambda_max}}
In order to prove \cref{prop:lambda_max}, we will use the following Lemma, which we prove separately in \cref{app:proof_Lemma_statistical}. 
\begin{lemma}\label{lemma:statistical}
Consider the setting $X \rightarrow Z \leftarrow Y$ as in \cref{subsec:solutions}. Assume the two statistical marginal models $\PP_{XZ}$, $\PP_{YZ}$ can successfully be merged, and $$\delta_X := \PP(Z=0 | X=1)- \PP(Z=0| X=0) \geq 0 \quad \text{and}\quad \delta_Y:= \PP(Z=0|Y=1) - \PP(Z=0|Y=0) \geq 0.$$ 
Then there exist 
conditional probabilities $q_{i,j} := \PP(Z=0|X=i, Y=j)$ such that $$q_{00} \leq q_{01} \leq q_{11} \quad \text{and} \quad q_{00}\leq q_{10} \leq q_{11}$$ and 
the distribution 
defined via $\PP_{XYZ}(X=i, Y=j, Z=0) = q_{ij}\PP(X=i) \PP(Y=j)$, has marginals that coincide with $\PP_{XZ}$ and $\PP_{YZ}$.
\end{lemma}
Analogous statements as in \cref{lemma:statistical} hold by swapping the roles of $X=1$ and $X=0$, or $Y=1$ and $Y=0$, respectively. For convenience we now restate \cref{prop:lambda_max} and then provide its proof.
\begin{prop_restated}
Consider the Boolean setting $X \rightarrow Z \leftarrow Y$ with marginal and joint models as defined in \cref{subsec:solutions}. If a solution to the structural causal marginal problem exists (i.e., $\Ccal$ is non-empty),
we  have $(\lamA^\text{max}, \lamB^\text{max})^\top \in \Lambda_{\mathcal{C}}$.
\end{prop_restated}

\begin{proof}[Proof of \cref{prop:lambda_max}]
As we do throughout we assume that
$ 0 <
\PP(X=1) <1$, and $0<
\PP(Y=1) < 1.$  

In the setting we consider and under the assumption that the statistical marginal models can be merged, there exists a conditional distribution  $\PP(Z|X,Y)$ such that  we have $\PP(X,Y,Z) = \PP(Z|X,Y) \PP(X) \PP(Y)$ and all statistical constraints are satisfied.
$\PP(Z|X,Y)$ is completely characterised by four probabilities $0\leq q_{i,j} \leq 1$ that are defined as $$q_{i,j} := \PP(Z=0|X=i, Y=j)$$ for $i,j\in \{0,1\}$. We will 
construct $\cb$ for the case that
\begin{align*}
    \textbf{(S1)}& \quad \PP(Z=0| X=0) \leq \PP(Z=0 | X=1) \quad \text{and} \quad \PP(Z=0|Y=0) \leq \PP(Z=0|Y=1).
\end{align*}
For all the other cases, the construction of $\cb$ follows in analogy.

By \cref{lemma:statistical}, there exists $ q_{00}, q_{01}, q_{10}, q_{11}$ that are consistent with $\PP_{XZ}$ and $\PP_{YZ}$ and such that $$q_{00} \leq q_{01} \leq q_{11} \quad \text{and} \quad q_{00}\leq q_{10} \leq q_{11}.$$  
Given $\{q_{ij}\}$, the question is, can we find a corresponding vector $\cb\in \Delta^{15}$ that implies $\lamA=\lamA^\text{max}$ and $\lamB = \lamB^\text{max}$?
The connection between $\{q_{ij}\}$ and $\cb$ is given in \cref{eq:causal_Markov_kernel} and can be derived from \cref{tab:15functions}:
\begin{align}\label{eq:c_to_q_gen}
\begin{aligned}
    q_{00} & = c_0 + c_2 +c_4 +c_6 + c_8 + c_{10} + c_{12} + c_{14}\\
    q_{01} & = c_0 + c_1 +c_4 +c_5 + c_8 + c_{9} + c_{12} + c_{13}\\
    q_{10} & = c_0 + c_1 +c_2 +c_3 + c_8 + c_{9} + c_{10} + c_{11}\\
    q_{11} & = c_0 + c_1 +c_2 +c_3 + c_4 + c_{5} + c_{6} + c_{7}
\end{aligned}
\end{align}

Recall that $(\Ab)_{jk}=\sum_{y=0}^1\PP_Y(y)\II\{\Pcal^Y_y(h_k)=f_j(X)\}$, with $f_0\equiv 0$, and $f_1\equiv1$, as well as $f_2(X)=X$ (``ID''), and $f_3(X)=1-X$~(``NOT''), see~\cref{sec:Boolean_cause_effect,sec:counterfactual_marginal_problem} and footnote 6.

By $\textbf{(S1)}$ setting $\lamA=\lamA^\text{max}$ results in $[\ab (\lamA)]_2 = 0 = [\Ab \cb]_2$. Writing out $[\Ab \cb]_2$ we obtain
\begin{align*}
    0 = c_4 \PP(Y=0) + c_6 \PP(Y=0) + c_8 \PP(Y=1) + c_9 \PP(Y=1) + c_{12} + c_{13}P(Y=1) + c_{14}P(Y=0).
\end{align*}
Together with $0 < \PP(Y=0) < 1$ and since all entries of $\cb$ are non-negative, this gives
$$0= c_4= c_6 = c_8 = c_9 = c_{12} = c_{13} = c_{14}.$$
Analogously, setting $\lamB=\lamB^\text{max}$ results in
$$0 = c_2= c_6 = c_8 = c_9 = c_{10} = c_{11} = c_{14}.$$

Thus, the only possible non-zero entries of $\cb$ are $c_0, c_1, c_3, c_5, c_7, c_{15}$ and \eqref{eq:c_to_q_gen} reduces to

\begin{align}\label{eq:c_to_q_spec}
\begin{aligned}
        q_{00} & = c_0 \\
    q_{01} & = c_0 + c_1 +c_5  \\
    q_{10} & = c_0 + c_1 + c_3 \\
    q_{11} & = c_0 + c_1  +c_3 + c_5 + c_{7}.
\end{aligned}
\end{align}
Since we have
$q_{00} \leq q_{01}, q_{10} \leq q_{11}$, we can make the following assignment:
\begin{align*}
    & \textbf{if $q_{01} \geq q_{10}$:}  \qquad & \textbf{if $q_{01}\leq q_{10}$:}\\
    &c_0 = q_{00} &c_0 = q_{00}\\
    & c_1 = q_{10} - q_{00} & c_1 = q_{01} - q_{00}\\
    & c_3 = 0 & c_3 = q_{10}-q_{01}\\
    & c_5 = q_{01} - q_{10} & c_5 =0\\
    & c_7 = q_{11} - q_{01} & c_7= q_{11}-q_{10}\\
    & c_{15} = 1 - q_{11} & c_{15} = 1- q_{11}
\end{align*}
Clearly the entries of $\cb$ sum to $1$ and since $q_{00} \leq q_{01} \leq q_{11}$ and $q_{00}\leq q_{10} \leq q_{11}$, all entries of $\cb$ are valid probabilities.
By construction we have now found a valid probability vector that is consistent with the marginals $\PP_{XZ}$ and $\PP_{YZ}$ and implies $\lamA = \lamA^\text{max}, \lamB=\lamB^\text{max}$.

Let us verify that the constructed $\cb$ indeed fulfils \eqref{eq:constraints_matrix} for $\lamA = \lamA^\text{max} = \PP(Z=0|X=0)$ (Recall that are working for the special case \textbf{(S1)}). Using the constructed $\cb$, \cref{tab:15functions}, the definition of $\Ab$ we obtain
\begingroup
\allowdisplaybreaks
\begin{align*}
    [\Ab\cb]_0 &= c_0 + c_1 \PP(Y=1) + c_5 \PP(Y=1) \\
    &= q_{00} + \PP(Y=1)(q_{01} - q_{00} )\\
    &= \PP(Y=0) q_{00} + \PP(Y=1)q_{01}\\
    & = \PP(Z=0, Y=0|X=0) + \PP(Z=0, Y=1|X=0)\\
    &= \PP(Z=0|X=0) = 0 + \lamA^\text{max} \\
    &= [\ab(\lamA^\text{max})]_0,\\[1em]
    [\Ab\cb]_1 &= c_5 \PP(Y=0) + c_7 \PP(Y=0) + c_{15} \\
    &= \PP(Y=0)(c_5 + c_7 +c_{15})+ \PP(Y=1) c_{15}\\
    &=\PP(Y=0)(q_{11}-q_{10} +1-q_{11}) +\PP(Y=1) (1-q_{11})\\
    &= 1 - \PP(Z=0,Y=0|X=1) - \PP(Z=0,Y=1|X=1)\\
    &= 1-\PP(Z=0|X=1) \\
    &= 1-\PP(Z=0|X=1)-\PP(Z=0|X=0) + \PP(Z=0|X=0) \\
    &= 1-\PP(Z=0|X=1)-\PP(Z=0|X=0) \lamA^\text{max}\\
    &= [\ab(\lamA^\text{max})]_1,\\[1em]
    [\Ab\cb]_2 &= 0\\
    & = \PP(Z=0|X=0) - \lamA^\text{max} \\
    &= [\ab(\lamA^\text{max})]_2,\\[1em]
    [\Ab\cb]_3 &=c_1 \PP(Y=0) + c_3 + c_7 \PP(Y=1) \\
    &= \PP(Y=0) (c_1 + c_3) + \PP(Y=1) (c_3 + c_7)\\
    &= \PP(Y=0) (q_{10}- q_{00}) + \PP(Y=1) (q_{11}-q_{01})\\
    &=\PP(Z=0, Y=0|X=1) - \PP(Z=0, Y=0|X=0) +  \PP(Z=0, Y=1|X=1) - \PP(Z=0, Y=1|X=0)\\
    &= \PP(Z=0|X=1) - \PP(Z=0|X=0)\\
    &= \PP(Z=0|X=1) - \lamA^\text{max}\\
    &= [\ab(\lamA^\text{max})]_3.
\end{align*}
\endgroup
Analogously, it follows that \eqref{eq:constraints_matrix} is consistent for $\lamB=\lamB^\text{max}$.
\end{proof}

\subsection{Proof of \cref{lemma:statistical}}\label{app:proof_Lemma_statistical}
\begin{proof}[Proof of \cref{lemma:statistical}]
    As we do throughout we assume that
$ <\PP(X=1) <1$, and $0<\PP(Y=1) < 1.$ 
    A necessary condition for merging $\PP_{XZ}$ and $\PP_{YZ}$ is that they imply the same marginal distribution  over $Z$
    \begin{align}\label{eq:pz0_consistent}
        \sum_{i\in\{0,1\}}\PP(Z=0|X=i) \PP(X=i) = \sum_{j\in\{0,1\}}\PP(Z=0|Y=j) \PP(Y=j).
    \end{align}
    
    Using the definition of $\delta_X, \delta_Y$ we obtain
    \begin{align}
        &\PP(Z=0|X=0)\PP(X=0) + (\PP(Z=0|X=0)+\delta_X) \PP(X=1) \\
        &\quad = (\PP(Z=0|Y=1) - \delta_Y)\PP(Y=0) + \PP(Z=0|Y=1) \PP(Y=1)\\
        \Leftrightarrow \qquad& \PP(Z=0|X=0) + \delta_X \PP(X=1) = \PP(Z=0|Y=1) - \delta_Y \PP(Y=0) \\
        \Leftrightarrow \qquad &\delta_{YX} := \PP(Z=0|Y=1) - \PP(Z=0|X=0) = \delta_Y \PP(Y=0) + \delta_X \PP(X=1).
    \end{align}
    By Assumption $\delta_X \geq 0, \delta_Y\geq 0$, and thus $\delta_{YX} \geq 0$. Analogously we obtain
    \begin{align}
        \delta_{XY} := \PP(Z=0|X=1) - \PP(Z=0|Y=0) = \delta_X \PP(X=0) + \delta_Y \PP(Y=1) \geq 0.
    \end{align}
    
    Since we assumed that a joint statistical model $\PP_{XYZ}$ exists, which is consistent with the marginals $\PP_{XZ},\PP_{YZ}$,
    there exists $q_{i,j} := \PP(Z=0|X=i, Y=j)$ such that
    \begin{align*}
        & \PP(Z=0|X=0) = q_{00} \PP(Y=0) + q_{01}\PP(Y=1),\\
        &\PP(Z=0|X=1) = q_{10} \PP(Y=0) + q_{11}\PP(Y=1),\\
        &\PP(Z=0|Y=0) = q_{00} \PP(X=0) + q_{10}\PP(X=1),\\
        &\PP(Z=0|Y=1) = q_{01} \PP(X=0) + q_{11}\PP(X=1).
    \end{align*}
    Thus under our assumption that none of the marginal probabilities $\PP(X=0)$, $\PP(Y=0)$ equals $0$ or $1$, choosing $q_{00}$ uniquely determines all the other $\{q_{ij}\}$:
    \begin{align}
    &\begin{aligned}
       (a)\quad & \PP(Z=0|X=0) = q_{00} \PP(Y=0) + q_{01}\PP(Y=1)\\
        & \Leftrightarrow q_{01} = \frac{\PP(Z=0|X=0)}{\PP(Y=1)} - \frac{q_{00}\PP(Y=0)}{\PP(Y=1)},
    \end{aligned}\\[1em]
        &\begin{aligned}
       (b)\quad & \PP(Z=0|Y=0) = q_{00} \PP(X=0) + q_{10}\PP(X=1)\\
        & \Leftrightarrow q_{10} = \frac{\PP(Z=0|Y=0)}{\PP(X=1)} - \frac{q_{00}\PP(X=0)}{\PP(X=1)},
    \end{aligned}\\[1em]
    &\begin{aligned}
       (c)\qquad & \PP(Z=0|Y=1) = q_{01} \PP(X=0) + q_{11}\PP(X=1)\\
         \Leftrightarrow q_{11} &= \frac{\PP(Z=0|Y=1)}{\PP(X=1)} - \frac{q_{01}\PP(X=0)}{\PP(X=1)}\\
         &=\frac{\PP(Z=0|Y=1)}{\PP(X=1)} - \frac{\left(\frac{\PP(Z=0|X=0)}{\PP(Y=1)} - \frac{q_{00}\PP(Y=0)}{\PP(Y=1)}\right)\PP(X=0)}{\PP(X=1)} \\
         &= \frac{\PP(Z=0|Y=1)}{\PP(X=1)} - \frac{\PP(Z=0|X=0)\PP(X=0)}{\PP(Y=1)\PP(X=1)} +  \frac{q_{00}\PP(Y=0)\PP(X=0)}{\PP(Y=1)\PP(X=1)}.
    \end{aligned}
    \end{align}
$\PP(Z=0|X=1) = q_{10} \PP(Y=0) + q_{11}\PP(Y=1)$ is then ensured if the marginals can consistently be merged, which we assumed.
Our goal is thus to check whether a $q_{00}$ exists such that 
\begin{align}
   0 \leq q_{00}, \quad q_{00} \leq q_{01}, \quad q_{00} \leq q_{10}, \quad q_{01} \leq q_{11}, \quad q_{10} \leq q_{11}, \quad q_{11}\leq 1.
\end{align}
Using the equalities defined above, we can express all these constraints in terms of $q_{00}$. Ensuring that a solution exists, will then complete the proof.

\begin{align*}
    (C1) \quad & 0 \leq q_{00}\\
    (C2) \quad & q_{00} \leq q_{01} \Leftrightarrow q_{00} \leq \frac{\PP(Z=0|X=0)}{\PP(Y=1)} - \frac{q_{00}\PP(Y=0)}{\PP(Y=1)} \Leftrightarrow q_{00} \leq \PP(Z=0|X=0),\\
    (C3) \quad & q_{00} \leq q_{10} \Leftrightarrow q_{00} \leq \frac{\PP(Z=0|Y=0)}{\PP(X=1)} - \frac{q_{00}\PP(X=0)}{\PP(X=1)} \Leftrightarrow q_{00} \leq \PP(Z=0|Y=0),\\
    (C4) \quad & q_{01} \leq q_{11} \Leftrightarrow  \frac{\PP(Z=0|X=0)}{\PP(Y=1)} - \frac{q_{00}\PP(Y=0)}{\PP(Y=1)} \\
    & \qquad\qquad\qquad\leq \frac{\PP(Z=0|Y=1)}{\PP(X=1)} - \frac{\PP(Z=0|X=0)\PP(X=0)}{\PP(Y=1)\PP(X=1)} +  \frac{q_{00}\PP(Y=0)\PP(X=0)}{\PP(Y=1)\PP(X=1)},\\
    (C5) \quad &  q_{10} \leq q_{11} \Leftrightarrow 
    \frac{\PP(Z=0|Y=0)}{\PP(X=1)} - \frac{q_{00}\PP(X=0)}{\PP(X=1)} \\
    & \qquad\qquad\qquad\leq 
    \frac{\PP(Z=0|Y=1)}{\PP(X=1)} - \frac{\PP(Z=0|X=0)\PP(X=0)}{\PP(Y=1)\PP(X=1)} +  \frac{q_{00}\PP(Y=0)\PP(X=0)}{\PP(Y=1)\PP(X=1)}\\
    (C6) \quad & q_{11} \leq 1 \Leftrightarrow 
     \frac{\PP(Z=0|Y=1)}{\PP(X=1)} - \frac{\PP(Z=0|X=0)\PP(X=0)}{\PP(Y=1)\PP(X=1)} +  \frac{q_{00}\PP(Y=0)\PP(X=0)}{\PP(Y=1)\PP(X=1)} \leq 1.
\end{align*}

$(C1),(C2), (C3)$ are already in interpretable form, so next we rewrite $(C4)$
\begingroup
\allowdisplaybreaks
\begin{align*}
    \,&\frac{\PP(Z=0|X=0)}{\PP(Y=1)} - \frac{q_{00}\PP(Y=0)}{\PP(Y=1)} \leq \frac{\PP(Z=0|Y=1)}{\PP(X=1)} - \frac{\PP(Z=0|X=0)\PP(X=0)}{\PP(Y=1)\PP(X=1)} +  \frac{q_{00}\PP(Y=0)\PP(X=0)}{\PP(Y=1)\PP(X=1)}\\
   \Leftrightarrow \,&{\PP(Z=0|X=0)} - {q_{00}\PP(Y=0)} \leq \frac{\PP(Z=0|Y=1)\PP(Y=1)}{\PP(X=1)} - \frac{\PP(Z=0|X=0)\PP(X=0)}{\PP(X=1)} \\
   \,& \qquad \qquad\qquad \qquad\qquad \qquad\qquad\quad+  \frac{q_{00}\PP(Y=0)\PP(X=0)}{\PP(X=1)}\\
   \Leftrightarrow \,&q_{00} \PP(Y=0) \left(1 + \frac{\PP(X=0)}{\PP(X=1)}\right) \geq 
   {\PP(Z=0|X=0)} 
   + \frac{\PP(Z=0|X=0)\PP(X=0)}{\PP(X=1)}  
   - \frac{\PP(Z=0|Y=1)\PP(Y=1)}{\PP(X=1)}\\
   \Leftrightarrow \,&q_{00} \PP(Y=0)  \geq 
   {\PP(Z=0|X=0)} \PP(X=1)
   + \PP(Z=0|X=0)\PP(X=0)
   - \PP(Z=0|Y=1)\PP(Y=1)\\
   \Leftrightarrow \,&q_{00} \PP(Y=0)  \geq 
   {\PP(Z=0|X=0)} 
   - \PP(Z=0|Y=1)\PP(Y=1)
   \\
   \Leftrightarrow \,&q_{00}   \geq \frac{
   {\PP(Z=0|X=0)} 
   - \PP(Z=0|Y=1)\PP(Y=1)}{\PP(Y=0)} \\
   \overset{(*)}{\Leftrightarrow} \,&q_{00}   \geq \frac{
   {\PP(Z=0|X=0)} 
   -\left[\PP(Z=0|X=0)\PP(X=0) + \PP(Z=0|X=1)\PP(X=1) - \PP(Z=0|Y=0) \PP(Y=0)\right]}{\PP(Y=0)}    \\
   \Leftrightarrow \,&q_{00}   \geq \PP(Z=0|Y=0) - \frac{\PP(Z=0|X=0)\PP(X=0) + \PP(Z=0|X=1)\PP(X=1) - \PP(Z=0|X=0)}{\PP(Y=0)}\\
   \Leftrightarrow \,&q_{00}   \geq \PP(Z=0|Y=0) - \frac{\PP(X=1)}{\PP(Y=0)}(\PP(Z=0|X=1) - \PP(Z=0|X=0))\\
   \Leftrightarrow \,&q_{00}   \geq \PP(Z=0|Y=0) - \frac{\PP(X=1)}{\PP(Y=0)}\delta_X, \qquad (C4)
\end{align*}
\endgroup
where at $(*)$ we used \eqref{eq:pz0_consistent}.
We can also rewrite $(C4)$ once more in terms of $\PP(Z=0|X=0)$
\begin{align*}
\,&q_{00}   \geq \PP(Z=0|Y=0) - \frac{\PP(X=1)}{\PP(Y=0)}\delta_X\\
\Leftrightarrow \,&q_{00}   \geq \PP(Z=0|Y=0) - \frac{\PP(X=1)}{\PP(Y=0)}(\PP(Z=0|X=1) - \PP(Z=0|X=0))\\
   \Leftrightarrow \,&q_{00}\PP(Y=0)   \geq \PP(Z=0|Y=0) \PP(Y=0) - \PP(X=1)(\PP(Z=0|X=1) - \PP(Z=0|X=0))
      \\
   \overset{(*)}{\Leftrightarrow} \,&q_{00}\PP(Y=0)   \geq \PP(Z=0|X=0)\PP(X=0) - \PP(Z=0|Y=1)\PP(Y=1)  +\PP(Z=0|X=0) \PP(X=1)
         \\
   {\Leftrightarrow} \,&q_{00}\PP(Y=0)   \geq \PP(Z=0|X=0) - \PP(Z=0|Y=1)\PP(Y=1)         \\
   {\Leftrightarrow} \,&q_{00}\PP(Y=0)   \geq \PP(Z=0|X=0)(\PP(Y=0)+\PP(Y=1))
   - \PP(Z=0|Y=1)\PP(Y=1) \\
   \,&\qquad \qquad \qquad - \PP(Z=0|X=0) \PP(Y=1) + \PP(Z=0|X=0) \PP(Y=1)
          \\
   {\Leftrightarrow} \,&q_{00}\PP(Y=0)   \geq  \PP(Z=0|X=0) \PP(Y=0) + \PP(Z=0|X=0)\PP(Y=1) - \PP(Z=0|Y=1)\PP(Y=1) 
    \\
   {\Leftrightarrow} \,&q_{00}   \geq  \PP(Z=0|X=0) -\delta_{YX}\frac{\PP(Y=1)}{\PP(Y=0)}. \qquad (C4)
\end{align*}

Analogously, we can work through $(C5)$ and obtain 
\begin{align*}
    &\frac{\PP(Z=0|Y=0)}{\PP(X=1)} - \frac{q_{00}\PP(X=0)}{\PP(X=1)} \leq 
    \frac{\PP(Z=0|Y=1)}{\PP(X=1)} - \frac{\PP(Z=0|X=0)\PP(X=0)}{\PP(Y=1)\PP(X=1)} +  \frac{q_{00}\PP(Y=0)\PP(X=0)}{\PP(Y=1)\PP(X=1)}\\
    \Leftrightarrow \quad & q_{00}\geq \PP(Z=0|X=0) - \delta_Y \frac{\PP(Y=1)}{\PP(X=0)} \\
    \Leftrightarrow \quad & q_{00}\geq \PP(Z=0|Y=0) - \delta_{XY} \frac{\PP(X=1)}{\PP(X=0)}. 
    \qquad (C5)
\end{align*}

Next we consider (C6)
\begin{align*}
    \quad & q_{11} \leq 1 \Leftrightarrow 
     \frac{\PP(Z=0|Y=1)}{\PP(X=1)} - \frac{\PP(Z=0|X=0)\PP(X=0)}{\PP(Y=1)\PP(X=1)} +  \frac{q_{00}\PP(Y=0)\PP(X=0)}{\PP(Y=1)\PP(X=1)} \leq 1\\
     \Leftrightarrow \quad &\PP(Z=0|Y=1)\PP(Y=1) - \PP(Z=0|X=0)\PP(X=0) +  q_{00}\PP(Y=0)\PP(X=0) \leq \PP(X=1)\PP(Y=1)\\
     \Leftrightarrow \quad &q_{00}  \leq \frac{\PP(X=1)\PP(Y=1) - \PP(Z=0|Y=1)\PP(Y=1) + \PP(Z=0|X=0)\PP(X=0)}{\PP(Y=0)\PP(X=0)}.
\end{align*}
Using 
\begin{align*}
    &\PP(X=1)\PP(Y=1) - \PP(Z=0|Y=1)\PP(Y=1) + \PP(Z=0|X=0)\PP(X=0) 
    \\
    &= \PP(X=1,Y=1) - \PP(Z=0, Y=1) + \PP(Z=0, X=0)     \\
    &=\left[\PP(Z=0, X=1, Y=1) + \PP(Z=1, X=1, Y=1)\right]\\
    &\qquad - \left[\PP(Z=0,X=0, Y=1) + \PP(Z=0,X=1,Y=1)\right] \\
    &\qquad + \left[\PP(Z=0, X=0, Y=0) + \PP(Z=0,X=0, Y=1)\right]\\
    &= \PP(Z=1, X=1, Y=1) + \PP(Z=0, X=0, Y=0)\geq 0,
\end{align*}
we obtain for $(C6)$
\begin{align}
    q_{00} \leq \frac{\PP(Z=1, X=1, Y=1) + \PP(Z=0, X=0, Y=0)}{\PP(X=0)\PP(Y=0)}. \qquad (C6)
\end{align}

Let us summarize all constraints once more:
\begin{align*}
    &(C1) \quad  q_{00} \geq 0 =:\eta_1\\
    &(C2) \quad q_{00} \leq \PP(Z=0|X=0)=:\eta_2,\\
    &(C3) \quad q_{00} \leq \PP(Z=0|Y=0)=:\eta_3,\\
    &(C4) \quad q_{00}   \geq \PP(Z=0|Y=0) - \frac{\PP(X=1)}{\PP(Y=0)}\delta_X = \PP(Z=0|X=0) -\delta_{YX}\frac{\PP(Y=1)}{\PP(Y=0)}=:\eta_4,\\
    &(C5) \quad q_{00}\geq \PP(Z=0|X=0) - \delta_Y \frac{\PP(Y=1)}{\PP(X=0)} = \PP(Z=0|Y=0) - \delta_{XY} \frac{\PP(X=1)}{\PP(X=0)}=:\eta_5,\\
    &(C6) \quad q_{00}  \leq \frac{\PP(Z=1, X=1, Y=1) + \PP(Z=0, X=0, Y=0)}{\PP(X=0)\PP(Y=0)} =: \eta_6.
\end{align*}
We now have to check whether all the lower bounds on $q_{00}$ are smaller than all the upper bounds, in other words:
\begin{align}
\text{The proof is complete}. \quad \Leftrightarrow \quad \max(\eta_1, \eta_4, \eta_5) \leq \min(\eta_2, \eta_3, \eta_6).
\end{align}
First consider the case where $ \eta_2 =\min(\eta_2, \eta_3, \eta_6)$. Then all lower bounds are achievable, since $\delta_X, \delta_Y, \delta_{XY}, \delta_{YX}$ are per definition non-negative. The same holds if $\eta_3 =\min(\eta_2, \eta_3, \eta_6)$.

It is less apparent to see what holds in the case of $\eta_6 =\min(\eta_2, \eta_3, \eta_6)$. Since the numerator of $\eta_6$ is the sum of two probabilities we always have $\eta_6 \geq 0 = \eta_1$. We therefore need to show $\eta_6 \geq \eta_4, \eta_6 \geq \eta_5$. 

\begin{align*}
    &\eta_4 \leq \eta_6\\
    \Leftrightarrow \quad & \PP(Z=0|Y=0) - \frac{\PP(X=1)}{\PP(Y=0)}\delta_X \leq \frac{\PP(Z=1, X=1, Y=1) + \PP(Z=0, X=0, Y=0)}{\PP(X=0)\PP(Y=0)}\\
    \Leftrightarrow \quad& \PP(Z=0,Y=0)\PP(X=0) - \PP(X=0)\PP(X=1)\left[\PP(Z=0|X=1)-\PP(Z=0|X=0)\right] \\
    &\qquad \leq \PP(Z=1, X=1, Y=1) + \PP(Z=0, X=0, Y=0)\\
        \Leftrightarrow \quad& \left[\PP(Z=0,X=0, Y=0)+ \PP(Z=0,X=1,Y=0)\right]\PP(X=0) - \PP(X=0)\PP(Z=0,X=1)\\
        &\qquad \qquad+\PP(X=1)\PP(Z=0,X=0) \\
    &\qquad \leq \PP(Z=1, X=1, Y=1) + \PP(Z=0, X=0, Y=0)\\        
    \Leftrightarrow \quad& \left[\PP(Z=0,X=0, Y=0)+ \PP(Z=0,X=1,Y=0)\right]\PP(X=0) - \PP(X=0)\PP(Z=0,X=1)\\
    &\qquad \qquad + \PP(X=1)\left[\PP(Z=0,X=0, Y=0) + \PP(Z=0,X=0,Y=1)\right] \\
    &\qquad \leq \PP(Z=1, X=1, Y=1) + \PP(Z=0, X=0, Y=0)\\
        \Leftrightarrow \quad& \PP(Z=0,X=1,Y=0)\PP(X=0) - \PP(X=0)\PP(Z=0,X=1) + \PP(X=1)\PP(Z=0,X=0,Y=1) \\
    &\qquad \leq \PP(Z=1, X=1, Y=1)\\
    \Leftrightarrow \quad& \PP(X=0)\left[\PP(Z=0,X=1,Y=0) - \PP(Z=0,X=1)\right] + \PP(X=1)\PP(Z=0,X=0,Y=1) \\
    &\qquad \leq \PP(Z=1, X=1, Y=1)\\
        \Leftrightarrow \quad& -\PP(X=0) \PP(Z=0,X=1,Y=1)  + \PP(X=1)\PP(Z=0,X=0,Y=1) \\
    &\qquad \leq \PP(Z=1, X=1, Y=1)\\
    \Leftrightarrow \quad&    \PP(X=1)\PP(Z=0,X=0,Y=1)  \leq \PP(Z=1, X=1, Y=1) + \PP(X=0) \PP(Z=0,X=1,Y=1)\\
    \Leftrightarrow \quad&    \PP(X=1)\PP(Z=0,X=0,Y=1)  \leq \PP(Z=1, X=1, Y=1) + (1-\PP(X=1)) \PP(Z=0,X=1,Y=1) \\
    \Leftrightarrow \quad&    \PP(X=1)\PP(Z=0,X=0,Y=1)  \leq \PP(X=1, Y=1) - \PP(X=1) \PP(Z=0,X=1,Y=1)\\
    \Leftrightarrow \quad&    \PP(X=1)\PP(Z=0,X=0,Y=1)  \leq \PP(X=1)\PP(Y=1) - \PP(X=1)\PP(Z=0,X=1,Y=1) \\
    \Leftrightarrow \quad&    \PP(X=1)\PP(Z=0,Y=1)  \leq \PP(X=1)\left[\PP(Z=0,Y=1) + \PP(Z=1, Y=1)\right]\\
    \Leftrightarrow \quad&   0  \leq \PP(X=1) \PP(Z=1, Y=1),
\end{align*}
which always is a true statement.
Analogously we obtain
\begin{align*}
    \eta_5 \leq \eta_6 \Leftrightarrow 0  \leq \PP(Y=1) \PP(Z=1, X=1) \Leftrightarrow \textbf{TRUE}.
\end{align*}
 Note that the validity of $\eta_6$ being larger than all lower bounds holds independently of our assumptions on $\delta_X\geq 0$ and $\delta_Y\geq 0$.

We have thus shown that under the assumptions stated in the Lemma, the interval $[\max(\eta_1, \eta_4, \eta_5) , \min(\eta_2, \eta_3, \eta_6)]$ is non-empty. Hence, picking any $q_{00}$ in this interval and computing $q_{01}, q_{10}, q_{11}$ accordingly, leads to an example fulfilling the Lemma's statement. 

\end{proof}

\section{Construction of the convex polytope $\mathcal{C}$}
\label{app:reformulation_linear_program}
In \cref{subsec:solutions} we defined the set of feasible joint counterfactual models via \cref{eq:C_polytope}. To clarify that this actually is a polytope and to handle it with numerical solvers, we need to express the constraints on $\cb$ as a set of equalities and inequalities. These should take the following form (see Section 2.2.4 of \citet{boyd2004convex}):
\begin{align}
    \Abt \cb &\preceq \bbt \label{ineq:eq_constr}\\
    \Cbt \cb &= \dbt \label{eq:eq_constr},
\end{align}
for some matrices $\Abt, \Cbt$ that need to be determined.
Here we take on the notation $\ab\preceq \bb$, to denote that $a_i \leq b_i$ for all entries.

\paragraph{Inequality constraints.} 
Starting from \eqref{eq:constraints_matrix} we have
\begin{align*}
    \Ab \cb &= \ab(\lambda^A) =
     \ab_0
    +
    \lambda^A
    \begin{pmatrix}
    1\\
    1\\
    -1\\
    -1
    \end{pmatrix}, \qquad \text{with}\qquad \ab_0 =
\begin{pmatrix}
    0\\
    1-P(Z=0|X=0)-P(Z=0|X=1)\\
    P(Z=0|X=0)\\
    P(Z=0|X=1)
    \end{pmatrix}\,.
    \end{align*}
Therefore, using $\lamA^\text{min} \leq \lambda^A \leq \lamA^\text{max}$, we get the following inequalities:
\begin{equation}
\Ab \cb \preceq \ab_0 + 
\begin{pmatrix}
\lamA^\text{max}\\
\lamA^\text{max}\\
-\lamA^\text{min}\\
-\lamA^\text{min}
\end{pmatrix}.
\label{eq:ineq_constr_1}
\end{equation}
And similarly for $\bb(\lambda^B)$,
\begin{equation}
\Bb \cb \preceq \bb_0 + 
\begin{pmatrix}
\lamB^\text{max}\\
\lamB^\text{max}\\
-\lamB^\text{min}\\
-\lamB^\text{min}
\end{pmatrix}.
\label{eq:ineq_constr_2}
\end{equation}
Additionally, from 
\begin{equation}
    -\Ab \cb = -\ab_0
    +
    \lambda^A
    \begin{pmatrix}
    -1\\
    -1\\
    +1\\
    +1
    \end{pmatrix}
\end{equation}
and again using $\lamA^\text{min} \leq \lambda^A \leq \lamA^\text{max}$, we get 
\begin{equation}
    -\Ab \cb \preceq -\ab_0
    +
    \begin{pmatrix}
    -\lamA^\text{min}\\
    -\lamA^\text{min}\\
    \lamA^\text{max}\\
    \lamA^\text{max}
    \end{pmatrix}.
    \label{eq:ineq_constr_3}
\end{equation}
And similarly for $\bb(\lambda^B)$,
\begin{equation}
-\Bb \cb \preceq -\bb_0 + 
\begin{pmatrix}
-\lamB^\text{min}\\
-\lamB^\text{min}\\
\lamB^\text{max}\\
\lamB^\text{max}
\end{pmatrix}.
\label{eq:ineq_constr_4}
\end{equation}

Finally, from the positivity constraint $c_i \geq 0 \,\,\, \forall i$, we get
$$
- \mathbb{I} \cb \leq [0, \ldots, 0]^\top \in \RR^{16}
$$
where $\mathbb{I}$ is the $16 \times 16$ identity matrix.

Overall, we can express the inequality constraints as
\begin{align}
    \Abt \cb \preceq \bbt
\end{align}
by defining the $32 \times 16$ matrix $\Abt$ and the $32$-dimensional vector $\bbt$ as
$$
    \Abt := \begin{pmatrix}
\Ab \\
- \Ab \\
\Bb \\
- \Bb \\
-\mathbb{I}
\end{pmatrix}, \qquad
    \bbt := \begin{pmatrix}
\abt_1 \\
\abt_2 \\
\bbt_1 \\
\bbt_2 \\
\mathbf{0}
\end{pmatrix}
$$
Where $\abt_1, \abt_2, \bbt_1, \bbt_2$ respectively denote the RHS of eqs.~\eqref{eq:ineq_constr_1},~\eqref{eq:ineq_constr_3},~\eqref{eq:ineq_constr_2},~\eqref{eq:ineq_constr_4}, and $\mathbf{0} = [0, \ldots, 0]^\top \in \RR^{16}$.

\paragraph{Equality constraints.}
Besides the inequalities above we also need to ensure that all 4 implied equations of $\ab(\lambda^A) = \Ab\cb$ are fulfilled simultaneously (i.e., they are fulfilled for the same $\lambda^A$). We can enforce this by ensuring that $\lambda^A$ as computed from the first row equals the one computed from the second, third, and fourth row, respectively. Let us make it explicit for the equality of $\lambda^A$ computed from the first two rows:
\begin{align*}
    & \lambda^A = [\Ab\cb]_0 - [\ab_0]_0 = [\Ab\cb]_1 - [\ab_0]_1 = \lambda^A\\
    &[\Ab\cb]_0 - [\Ab\cb]_1 = [\ab_0]_0 - [\ab_0]_1\\
    \Leftrightarrow 
    &\begin{pmatrix}
    1 & -1 & 0 & 0
\end{pmatrix}
\Ab \cb = \begin{pmatrix}
    1 & -1 & 0 & 0
\end{pmatrix}
\ab_0
\end{align*}

Doing this also for the third and fourth row, we obtain the constraints

\begin{align*}
    \begin{pmatrix}
    1 & -1 & 0 & 0
\end{pmatrix}
&\Ab \cb = \begin{pmatrix}
    1 & -1 & 0 & 0
\end{pmatrix}
\ab_0,\\
    \begin{pmatrix}
    1 & 0 & 1 & 0
\end{pmatrix}
&\Ab \cb = \begin{pmatrix}
    1 & 0 & 1 & 0
\end{pmatrix}
\ab_0,\\
    \begin{pmatrix}
    1 & 0 & 0 & 1
\end{pmatrix}
&\Ab \cb = \begin{pmatrix}
    1 & 0 & 0 & 1
\end{pmatrix}
\ab_0,
\end{align*}
which we rewrite as one set of constraints
\begin{align}
\Cbt_\Ab \cb =         \dbt_{\ab_0}, \quad \text{with  }        
{\Cbt_\Ab}:= \begin{pmatrix}
        1 & -1 & 0 & 0 \\
        1 & 0 & 1 & 0\\
        1 & 0 & 0 & 1
\end{pmatrix}
&\Ab, \quad \text{and } 
\dbt_{\ab_0} :=\begin{pmatrix}
        1 & -1 & 0 & 0 \\
        1 & 0 & 1 & 0\\
        1 & 0 & 0 & 1
\end{pmatrix}
\ab_0.
\end{align}
Proceeding similarly, we obtain
\begin{align}
\Cbt_\Bb \cb =         \dbt_{\bb_0}, \quad \text{with  }        
{\Cbt_\Bb}:= \begin{pmatrix}
        1 & -1 & 0 & 0 \\
        1 & 0 & 1 & 0\\
        1 & 0 & 0 & 1
\end{pmatrix}
&\Bb, \quad \text{and } 
\dbt_{\bb_0} :=\begin{pmatrix}
        1 & -1 & 0 & 0 \\
        1 & 0 & 1 & 0\\
        1 & 0 & 0 & 1
\end{pmatrix}
\bb_0.
\end{align}

Additionally we obtain one equality constraint ensuring that the probabilities of $\cb$ sum to one $\sum_i c_i = 1$, i.e.,
$$
\Cbt_{\bm{1}} \cb =1, \quad \text{with }
\Cbt_{\bm{1}} := [1.0, \ldots, 1.0]\,, \RR^{1\times 16}\,.
$$
Overall we can thus collect all equality constraints as $\Cbt \cb = \dbt$ with 
\begin{align*}
    \Cbt := \begin{pmatrix}
        \Cbt_{\Ab}\\
        \Cbt_{\Bb} \\
        \Cbt_{\bm{1}}
    \end{pmatrix},\qquad
    \text{and } 
    \dbt := \begin{pmatrix}
        \dbt_{\ab_0}\\
        \dbt_{\bb_0}\\
        1
    \end{pmatrix}
\end{align*}

\paragraph{Characterization of $\mathcal{C}$.} To summarise we can characterise the polytope of feasible joint models as
\begin{align*}
\boxed{
    \mathcal{C}:= \{\cb \in \Delta^{15} \mid 
    \exists (\lamA, \lamB) \in \Lambda_0 \,\,\,
    \text{s.t.} \,\,\, \eqref{eq:constraints_matrix} \,\,\, \text{holds}
    \} = \{\cb\in \mathbb{R}^{16} \mid \Cbt \cb = \dbt, \Abt \cb = \bbt  \}
    }.
\end{align*}

The convex polyhedron $\mathcal{C}$ can alternatively be represented as the convex hull of its vertices $\mathcal{V}_\mathcal{C}:=\{\bm{v}_1,\ldots,\bm{v}_m \}\subset\mathbb{R}^{16}$, where $m\in\mathbb{N}$ depends on the number of constraints.
We use the \texttt{pypoman} package~\cite{caron2018}, to compute those vertices of the polyhedron. 
After that we can project each vertex into the $(\lamA,\lamB)$-plane. 
Since $[\Ab \cb]_0 =\lamA$ and $[\Bb \cb]_0 =\lamB$, each $\cb$ corresponds to 
\begin{align}
    \begin{pmatrix}
        \lamA\\
        \lamB
    \end{pmatrix}
    = \Ebt \cb,
\end{align}
with
\begin{align}
    \Ebt := \begin{pmatrix}
    1 & 0 & 0 & 0 & 0 & 0 & 0 & 0\\
    0 & 0 & 0 & 0 & 1 & 0 & 0 & 0\\
\end{pmatrix} \cdot 
\begin{pmatrix}
    \Ab \\
    \Bb\\
\end{pmatrix}.
\end{align}
Thus the region of admissible pairs $(\lamA,\lamB)$ (green region in \cref{fig:overview}) is given as the convex hull of the projected vertices of $\mathcal{C}$
\begin{align}
    \Lambda_\mathcal{C} = \text{Conv}\left(\{\Ebt \bm{v} \mid \bm{v} \in \mathcal{V}_\mathcal{C}\}  \right).
\end{align}

\section{Technical details on the Experiments}
\label{app:experiments}
All the code used for the experiments and to generate the plots in~\cref{fig:experiments} can be found in the supplementary material.

\textbf{Parametrising $\PP_{XYZ}$.} In order to ensure that a solution to the marginal problem exists, we start from parameters of the joint distribution $\PP_{XYZ}$ (i.e. a joint distribution exists by construction), and compute the parameters of the marginal distributions $\PP_X$ and $\PP_Y$ by marginalisation.
We factorise %
$\PP_{XYZ}$ according to the DAG in~\cref{fig:teaser} as 
\begin{align}
    \theta_{X}&=P(X=1) \label{eq:marg_x}\\
    \theta_{Y}&=P(Y=1) \label{eq:marg_y}\\
    \theta_{Z|X=x, Y=y}&=P(Z=1|X=x, Y=y), \,\,\, x, y \in \{ 0 , 1\} \label{eq:cond_z_xy}\,,
\end{align}
which overall requires $6$ parameters.

\textbf{Finding solutions: projections of the $16$-dimensional polytope.} 
The polygon $\Lambda_\Ccal$ can be determined by projecting the vertices of the high-dimensional polytope $\Ccal$ in $2$-dimensions and computing their convex hull.
For the polytope vertices projection, the equality and inequality constraints and affine projection %
are described in~\cref{app:reformulation_linear_program}.
For polyhedra manipulation in Python, we use the \texttt{pypoman} package~\cite{caron2018}, which allows to compute the $2$-d projection of the vertices of our $\Ccal$ polytope.
Once the projected vertices are computed, $\Lambda_\Ccal$ can be found as their convex hull, which we compute using \texttt{scipy}~\cite{virtanen2020scipy}.

Alternatively, if we are only interested in  $\LBA^\star, \UBA^\star$%
, the computation %
could be formulated as a linear program~\cite{boyd2004convex},
\begin{equation}
    \begin{aligned}
    \minmax_{\lamA,\lamB\in\RR, \cb\in\Delta^{15}} \quad  & \lamA \\
    \text{subject to} \quad  & \ab(\lamA)= \Ab\cb\\
    & \ab(\lamB)= \Bb\cb\\
    & \lamA^\text{min}\leq\lamA\leq\lamA^\text{max}\\
    & \lamB^\text{min}\leq\lamB\leq\lamB^\text{max}
    \end{aligned}
    \label{eq:old_lp_formulation}
\end{equation}
and similarly for $\LBB^\star, \UBB^\star$.
This can be solved
using e.g. the \texttt{linprog} module in \texttt{scipy}~\cite{virtanen2020scipy}.

\textbf{Sampling problem instances.} 
The parameters in~\eqref{eq:marg_x} and~\eqref{eq:marg_y} are sampled from a Uniform distribution on $[0, 1]$, while those in~\eqref{eq:cond_z_xy} are sampled from a Beta distribution, whose parameters $\alpha$ and $\beta$ are set either to $1$ (corresponding to a Uniform distribution) or to $0.5$ (which puts more mass towards the extremes, thus resulting in more deterministic conditional distributions $\PP_{Z|XY}$).

\textbf{Parameter sweeps and GIF visualisations.} To generate~\href{https://ibb.co/rf9XYzD}{[\underline{GIF1}]} we use generic (i.e., not inducing a unique SCM) conditionals
\begin{align}
&\PP(Z=1 | X=0, Y=0) = 0.3,\\
&\PP(Z=1 | X=0, Y=1) = 0.8,\\
&\PP(Z=1 | X=1, Y=0) = 0.3,\\
&\PP(Z=1 | X=1, Y=1) = 0.7.
\end{align}

Whereas~\href{https://ibb.co/nB3dTrg}{[\underline{GIF2}]} is generated through a joint SCM $Z:=X \oplus Y$:
\begin{align}
&\PP(Z=1 | X=0, Y=0) = 0,\\
&\PP(Z=1 | X=0, Y=1) = 1,\\
&\PP(Z=1 | X=1, Y=0) = 1,\\
&\PP(Z=1 | X=1, Y=1) = 0.
\end{align}

In both cases we then sweep over different probabilities $\PP(X=1), \PP(Y=1)$ as shown in the respective plots on the right.

The green points in the plots represent the projected vertices of the high-dimensional polytope $\Ccal$.

\section{An illustrative example}
\label{app:illustrative_example}
We now provide an example which---despite arguably being slightly contrived%
---is meant to illustrate the potential usefulness of our approach and the structural causal marginal problem in a real-world context. 

Suppose that we are interested in investigating a disease $Z$ for which $Z=1$  indicates that a person recovers completely after ten days (fast recovery), while $Z=0$ indicates that the disease went on for more than ten days (long symptoms). We assume that there exists some medication $X$ against disease $Z$, such that $X=1$ denotes that a person took the medication, while $X=0$ denotes that a person did not take the medication.

Clearly, the disease does not cause the medication, but potentially vice versa, so we can take the causal graph to be $X\to Z$.
For sake of simplicity, suppose further that $X$ and $Z$ are unconfounded (see~\cref{app:confounded_case} for a detailed treatment of confounding). 

We have access to an observational study in the form of  a distribution $\PP_{XZ}$ which indicates that \textit{without} medication the chances of having long symptoms are $50\%$, i.e., $\PP(Z=0|X=0) = 1/2$, whereas \textit{with} medication the chances of long symptoms reduce to $40\%$, i.e., $\PP(Z=0|X=1) = 0.4$.
Thus, overall, the medication has a positive ACE. 

The family of marginal SCMs $\MA$ over $X\to Z$ that can explain these findings can be found via \eqref{eq:lam_A} and are given by:
\begin{equation}\label{eq:illustrative_range_a}
    \ab(\lamA)
=
\begin{pmatrix}
0\\
0.1\\
0.5\\
0.4
\end{pmatrix}
+\lamA
\begin{pmatrix}
1\\
1\\
-1\\
-1
\end{pmatrix},
\end{equation}
with $\lamA\in [0, 0.4]$. 

Since $\lamA=0$ is allowed, we cannot exclude that what happens is the following:
\begin{itemize}[leftmargin=3em]
    \item For $10\%$ of the people the medication has no effect and they always recover fast ($a_1 = 0.1$, $Z:=f_1(X)\equiv1$).
    \item For $50\%$ of the people the medication causes the fast recovery, while without medication, they have long symptoms ($a_2 = 0.5$, $Z:=f_2(X)=X$).
    \item For $40\%$ of the people the medication causes active harm: If they take it, they experience long symptoms, while without, they recover fast ($a_3 =0.4$, $Z:=f_3(X)=1-X$).
\end{itemize}

Now it is plausible that this scenario would be quite frightening and could cause some people to refuse to take the medication because they are afraid that it harms. (Although from a purely statistical perspective it is still advisable to take it and that's why we assume the medication was approved.)

But now imagine that another study is conducted that investigates the (unconfounded) effect of the presence of some specific genotype $Y=1$ ($Y=0$ denotes that a person has a different genotype than the one under investigation) on the chances of fast recovery $Z=1$. For, say, privacy reasons, however, \emph{this study does not document whether or not subjects undertook the medication} $X$, so it only provides data from $\PP_{YZ}$ and we have no joint observations of $\PP_{XYZ}$.

Suppose the second study finds that $40\%$ of  people have the genotype, $\PP(Y=1)=0.4$, and \emph{all} of those experience long symptoms $\PP(Z=0|Y=1)=1$.

We fix the remaining probabilities to $\PP(X=0)=\PP(X=1)=1/2$ and $\PP(Z=0|Y=0) = 1/12$, although other choices can also lead to the same conclusion.

With the methods proposed in this paper, we can then show that enforcing consistency of both datasets constrains the possible SCMs over $X\rightarrow Z$ to a \emph{unique} $\lamA = 0.4$, see~\cref{app:calculation_example} below for details. 
Now this SCM has a totally different interpretation to the one (previously still possible) given above:
\begin{itemize}[leftmargin=3em]
    \item For $40\%$ of people the medication has no effect and they always experience long symptoms ($a_0 = 0.4$, $Z:=f_0(X)\equiv0$).
    \item For $50\%$ of people the medication has no effect and they always recover fast ($a_1 = 0.5$, $Z:=f_1(X)\equiv 1$).
    \item For $10\%$ of people the medication causes the fast recovery, while without medication, they have long symptoms ($a_2 = 0.1$, $Z:=f_2(X)=X$).
    \item For $0\%$ of people the medication causes active harm ($a_3 =0$, $Z:=f_3(X)=1-X$).
\end{itemize}

It seems plausible that people would be much more willing to take the medication now that they know '\emph{it cannot harm}'---even if the ACE remains unchanged. However, note that we now also know that the medication only helps in $10\%$ of the cases.

\subsection{Explicit Calculation}\label{app:calculation_example}
For conciseness when we presented the example we simply stated that $\PP_{YZ}$ forces $\lamA=0.4$. For completeness we now provide the explicit calculation.
We assumed $\PP(Z=0|Y=1)=1$, thus from \eqref{eq:lam_A} we obtain
\begin{align}
        \bb(\lamB)
=
\begin{pmatrix}
0\\
-\PP(Z=0|Y=0)\\
\PP(Z=0|Y=0)\\
1
\end{pmatrix}
+\lamB
\begin{pmatrix}
1\\
1\\
-1\\
-1
\end{pmatrix}.
\end{align}
The only value of $\lamB$ that ensures this is a valid probability vector is $\lamB = \PP(Z=0|Y=0)$. This results in 
\begin{align}
        \bb
=
\begin{pmatrix}
\PP(Z=0|Y=0)\\
0\\
0\\
1-\PP(Z=0|Y=0)
\end{pmatrix}.
\end{align}
By enforcing $\Bb \cb = \bb$, for the two zero entries we obtain (by similar considerations as in the proof of \cref{prop:lambda_max}):
\begin{align*}
    &0 = [\Bb \cb]_1 = c_3\PP(X=0) + c_7 \PP(X=0) + c_{11}\PP(X=0) +c_{12}\PP(X=1) + c_{13}\PP(X=1) + c_{14}\PP(X=1) + c_{15}\\
    &\Leftrightarrow 0 = c_3 = c_7 = c_{11} = c_{12} = c_{13} = c_{14} =  c_{15}.
\end{align*}
Similarly 
\begin{align*}
    &0 = [\Bb \cb]_2 = c_2 \PP(X=0) + c_6 \PP(X=0) + c_{10} +c_{14}\PP(Y=0) + c_8 \PP(X=1) + c_9\PP(X=1) + c_{11}\PP(X=1)\\
    &\Leftrightarrow 0 = c_2 = c_6 = c_{10} = c_{14} =  c_{8} = c_{9} =  c_{11}.
\end{align*}
So overall the non-zero entries can only be $c_0, c_1, c_4, c_5$

Furthermore, we have (considering only non-zero entries of $\cb$)
\begin{align}
    [\Ab\cb]_0 &= c_0 + c_1\PP(Y=1) + c_4\PP(Y=1) + c_5 \PP(Y=1)\\
        &= \PP(Y=0)c_0 + \PP(Y=1)(c_0+c_1+c_4+c_5)\\
        &=0.6 c_0 + 0.4 \geq 0.4
\end{align}
where we used $\PP(Y=1)=0.4$ and $c_0+c_1+c_4+c_5=1$ as they are the only non-zero entries. On the other hand from \eqref{eq:illustrative_range_a} we have $[\ab(\lamA)]_0 \leq 0.4$. Hence the only way $[\Ab\cb]_0 = [\ab(\lamA)]_0=\lamA$, happens if $c_0 = 0$ and $\lamA = 0.4$.

To conclude, we show that setting $c_1=0$, $c_4 = \frac{1}{6}$, $c_5 = \frac{5}{6}$ leads to the correct marginals. 

We have
\begin{align}
    \PP(Z=0|X=0) &= c_4 + \PP(Y=1)c_5 = \frac{1}{6} + \frac{5}{6} \cdot 0.4 = 1/2 \quad \checkmark\\
    \PP(Z=0|X=1) &= c_4 \PP(Y=1) + c_5 \PP(Y=1) = (c_4+c_5) \cdot \PP(Y=1) =   0.4 \quad \checkmark\\
    \PP(Z=0|Y=0) &= c_4 \PP(X=0) = \frac{1}{6}\cdot\frac{1}{2} = \frac{1}{12} \quad \checkmark\\
    \PP(Z=0|Y=1) &= c_4 +c_5 = 1 \qquad \checkmark.
\end{align}
Furthermore we have 
\begin{align*}
    &\PP(Z=0|X=0) \PP(X=0) + \PP(Z=0|X=1)\PP(X=1)= \frac{1}{2} (0.5 + 0.4) \\
&= 0.45 \\
&= \frac{1}{12}\cdot\frac{3}{5} + 1 \cdot \frac{4}{10} = \PP(Z=0|Y=0)\PP(Y=0) + \PP(Z=0|Y=1)\PP(Y=1),
\end{align*}
so also the marginal distribution over $Z$ is consistent. 

\section{More on the connection to statistical learning theory and capacity measures}
\label{app:learning_theory}
The following remarks are meant to illustrate to what extent the ambiguity in the space of allowed SCMs may be reduced by only considering function classes with low VC dimension~\cite{vapnik1971uniform}.
Intuitively, if we allow all (arbitrarily complex) response functions, the space of consistent (joint) models can be quite large. If, on the other hand, we constrain their allowed capacity and only allow for `simple' functions, this couples their behaviour across different input values and consequently can reduce the model space substantially.

Since the space of possible SCMs compatible with all observed probabilities is a convex polytope, a simple measure for its size is the entropy of its unique maximum entropy distribution. 
Let us first compute this entropy for the case where all response functions are allowed (i.e., without restrictions on their VC dimension).

To generate any conditional $\PP_{Y|X}$ %
with cause $X$ and effect $Y$ attaining values in finite sets $\cal{X}$ and $\cal{Y}$ with $|\Xcal|=n$ by an SCM, following~\citet[][\S~3.4]{peters2017elements} we represent each function $f:{\cal X} \to {\cal Y}$ as  an
element in the $n$-fold Cartesian product
${\cal Y} ^n := \cal{Y}\times \cdots \times \cal{Y}$ such that the $j$-th component indicates $f(x_j)$.
Then, each distribution $\PP(Y|X=x_j)$ determines only the marginal distribution of the $j$\textsuperscript{th} component of~$\Ycal^n$.  
Thus, the MaxEnt joint distribution on ${\cal Y}^n$
having these $n$ marginal distributions is simply given by their product. In other words, we obtain a distribution of functions 
in which observing what $f$ does with the input $x_j$ tells us nothing on what $f$ does with a different input $x_i\neq x_j$.
The observed probabilities are always compatible with such a 'decoupling of inputs' since we don't observe one draw of the function applied to different inputs. 
The SCM obtained this way has the entropy 
\begin{equation}\label{eq:sument}  
\sum_{j=1}^n H(Y|X=x_j),
\end{equation} 
where the sum runs over all $n$ possible values $x_j$ of $X$ (without weighting factor $p(x_j)$) and thus grows as $O(n)$.

Restricted function classes, however, couple different inputs:
If $Y$ is binary, and we consider a function class $C$ with VC dimension $h$, the size $|C|$ (which here coincides with the shattering coefficient) is bounded from above by $\log |C|\in O(h \log n)$ \cite{Vapnik}. Hence, for fixed $h$, the MaxEnt distribution on $C$ grows at most logarithmically in $n$ as opposed to the linear growth in \eqref{eq:sument}. 

In summary, this means that the space of allowed models (as measured by MaxEnt here), is reduced when restrictions on the function class are enforced.

\vfill

\end{document}